\documentclass[11pt]{article}

\usepackage{acl}

\usepackage{amsmath}
\usepackage{amssymb}
\usepackage{booktabs}
\usepackage{xspace}
\usepackage{pgfplots}
\pgfplotsset{compat=1.18}
\usepgfplotslibrary{groupplots,fillbetween}
\usepackage{subcaption}
\usepackage{multirow}
\usepackage[ruled,vlined,noend]{algorithm2e} 
\usepackage{enumitem}
\usepackage{wrapfig2}
\usetikzlibrary{positioning}
\usepackage{setspace}
\usepackage{makecell}
\usepackage{pifont}

\usepackage[capitalise]{cleveref}

\crefformat{section}{\S#2#1#3}
\Crefformat{section}{\S#2#1#3}
\crefrangeformat{section}{\S\S#3#1#4--#5#2#6}
\Crefrangeformat{section}{\S\S#3#1#4--#5#2#6}
\crefmultiformat{section}{\S\S#2#1#3}{ and~#2#1#3}{, #2#1#3}{, and~#2#1#3}
\Crefmultiformat{section}{\S\S#2#1#3}{ and~#2#1#3}{, #2#1#3}{, and~#2#1#3}

\crefformat{figure}{Fig.~#2#1#3}
\Crefformat{figure}{Fig.~#2#1#3}
\crefrangeformat{figure}{Figs.~#3#1#4--#5#2#6}
\Crefrangeformat{figure}{Figs.~#3#1#4--#5#2#6}
\crefmultiformat{figure}{Figs.~#2#1#3}{ and~#2#1#3}{, #2#1#3}{, and~#2#1#3}
\Crefmultiformat{figure}{Figs.~#2#1#3}{ and~#2#1#3}{, #2#1#3}{, and~#2#1#3}

\crefformat{table}{Tab.~#2#1#3}
\Crefformat{table}{Tab.~#2#1#3}
\crefrangeformat{table}{Tabs.~#3#1#4--#5#2#6}
\Crefrangeformat{table}{Tabs.~#3#1#4--#5#2#6}
\crefmultiformat{table}{Tabs.~#2#1#3}{ and~#2#1#3}{, #2#1#3}{, and~#2#1#3}
\Crefmultiformat{table}{Tabs.~#2#1#3}{ and~#2#1#3}{, #2#1#3}{, and~#2#1#3}

\newcommand{\blue}[1]{{#1}}

\usepackage[normalem]{ulem} 
\newcommand{\system}{\textsc{Orthrus}\xspace}
\newcommand{\dedicated}{Dedicated\xspace}
\newcommand{\sharedhomo}{Unified-Homo\xspace}
\newcommand{\sharedhetero}{Unified-Hetero\xspace}
\newcommand{\disaggpd}{Disagg-PD\xspace}
\newcommand{\fcfs}{FCFS\xspace}
\newcommand{\ibs}{IBS\xspace}
\newcommand{\ibsfull}{Intra-Batch Scheduling\xspace}
\newcommand{\code}{https://github.com/illinoisdata/Orthrus}

\definecolor{red}{HTML}{e21d23}
\definecolor{orange}{HTML}{E66100}
\definecolor{yellow}{HTML}{FFC107}
\definecolor{green}{HTML}{004D40}
\definecolor{blue}{HTML}{1E88E5}
\definecolor{navy}{HTML}{002873}
\definecolor{purple}{HTML}{816dc1}
\definecolor{phaseone}{HTML}{FFE9CC}
\definecolor{phasetwo}{HTML}{D6EFEA}
\definecolor{phasethree}{HTML}{E7DFF8}
\colorlet{generationcolor}{blue}
\colorlet{embeddingcolor}{red}
\colorlet{combinedcolor}{orange}

\title{Efficient Iterative Retrieval with Heterogeneous Batching}

\author{
	\textbf{Dohyun Park\textsuperscript{1}}, \quad
	\textbf{Hubertus Franke\textsuperscript{2}}, \quad
	\textbf{Daniel G. Waddington\textsuperscript{2}}, \\
	\textbf{Swaminathan Sundararaman\textsuperscript{2}}, \quad
	\textbf{Yongjoo Park\textsuperscript{1}} \\
	\textsuperscript{1}University of Illinois Urbana-Champaign \quad
	\textsuperscript{2}IBM Research \\
	{\texttt{\symbol{123}dohyunp2,yongjoo\symbol{125}@illinois.edu}}
	{\texttt{\symbol{123}frankeh,daniel.waddington,swami\symbol{125}@ibm.com}}}

\begin{document}

\maketitle

\begin{abstract}
Modern information retrieval increasingly employs both embedding and generative models to handle complex queries. However, current serving systems suffer from low throughput and poor GPU utilization because they execute these models in isolation. Coarse-grained partitioning, such as dedicating GPUs to specific tasks, fails to adapt to dynamic workloads and creates computational ``bubbles''.
To address these, we present \system, a serving system that performs \emph{heterogeneous batching} within a unified inference loop. The primary challenge lies in unifying embedding and generation workloads with conflicting computational patterns while optimizing batch composition for high performance. \system addresses these challenges through chunked embedding with incremental pooling and by adjusting batch composition in a workload-aware manner. Evaluation on four A100 GPUs shows that, relative to baseline deployments, \system achieves 1.28$\times$--4.52$\times$ higher throughput on controlled workloads and up to 55.8\% lower end-to-end p99 latency on an iterative-RAG benchmark.
We release our code at
\url{\code}.
\end{abstract}

\input{figures/fig_intro2}

\section{Introduction }

Modern retrieval techniques~\citep{asai_self-rag_2023, gao_precise_2023, jiang_active_2023, shao_enhancing_2023} integrate both embedding and generation models to identify information gaps and retrieve documents iteratively. Consider a user query: ``Does the Gold plan cover my gym membership?'' This query is first \emph{embedded} into a vector~\cite{wang_improving_2024,izacard2021contriever} to search a vector database~\cite{hnsw_2020,sieve_2025}. If a retrieved document states that ``The Gold plan covers Category B activities,'' the system clarifies the ambiguity by letting a \emph{generative} model~\cite{asai_self-rag_2023,shao_enhancing_2023} formulate a follow-up query---``What are Category B activities?''---which is then embedded to retrieve supplementary documents. This cycle continues until sufficient context has been gathered. To accelerate this, serving systems must be explicitly designed to handle mixed embedding/generation workloads.


Unfortunately, existing systems suffer from low throughput and poor GPU utilization when handling mixed workloads. Serving frameworks like vLLM~\cite{kwon_efficient_2023} are designed to execute embedding and generation models in isolation.
Simply running two serving instances (i.e., OS processes) on the same GPU---one for each task, as in \cref{fig:intro2}(b)---is inefficient in current architectures~\cite{yu2022orca}. In a multi-GPU setup, resources can be statically partitioned for embedding and generation (\cref{fig:intro2}(a)), but this approach creates two major issues: (i) accurately predicting the optimal resource ratio is difficult, and (ii) dynamic repartitioning requires costly model reloading.
Even with perfect workload foresight, coarse-grained GPU partitioning remains suboptimal because embedding tasks are typically compute-bound, whereas generation is memory-bound~\cite{cacheCraft_2025}.
We hypothesize that \emph{homogeneous batching} is a fundamental bottleneck in iterative retrieval.

\paragraph{Core Idea}

We present \system, a serving system optimized for mixed embedding and generation workloads. By unifying these distinct compute patterns into a shared abstraction, \system enables \textbf{\emph{heterogeneous batching}} by executing embedding and generation requests together in a shared model execution pass. Unlike homogeneous batching (\cref{fig:intro2}(b)), this approach co-schedules tasks to maximize resource utilization (\cref{fig:intro2}(c)) while eliminating the overhead of concurrent server instances. This architecture enables fine-grained load balancing and generalizes to multi-GPU settings, where every node can serve mixed workloads. Importantly, \system requires no additional model training, making it applicable to existing deployments.

\paragraph{Challenge}

We address two primary challenges.
\textbf{\emph{(Heterogeneous Workloads)}} To maximize efficiency, it is desirable to execute embedding and generation requests together in the same model execution pass. Forming such heterogeneous passes is non-trivial because the requests exhibit conflicting computational patterns. A generation decode step requires relatively little compute to predict the next token, whereas embedding requires substantial compute to aggregate internal states across a full input. Na\"ively batching them without chunking forces lightweight work to wait for heavier requests, causing \emph{bubbles}.
\textbf{\emph{(Batch Composition)}} The ratio of embedding to generation requests in a batch heavily impacts user-perceived performance, yet finding the optimal ratio is difficult. FCFS scheduling causes head-of-line blocking for specific request types. Static heuristics (e.g., a 1:2 ratio) are also brittle because the number of tokens generated varies per request, so the optimal resource mix shifts dynamically and makes fixed ratios suboptimal.


\paragraph{Our Approach}

We employ the following techniques.
\textbf{\textit{(Chunked Embedding for Compute Balance)}}
To balance embedding and generation work within a batch, we adapt incremental computation~\cite{deepola,onlineagg} to embedding inference. We call this execution strategy \emph{chunked embedding}: the input is processed in fine-grained token chunks across scheduling iterations, while \emph{incremental pooling} maintains the cross-chunk state needed to produce the final embedding (\cref{sec:batch:pooler}). 
The unified runner schedules embedding chunks alongside generation decode steps to balance compute loads and minimize \emph{bubbles}.

\textbf{\textit{(Intra-Batch Scheduling)}}
Our design is governed by a central principle: \emph{retrieval ends only when all embedding and generation requests are fully processed.} To achieve this, \system dynamically adjusts batch composition using a \emph{cost-aware queue length}. The scheduler measures each backlog by its total remaining token count rather than its raw request count, allocating capacity in proportion to the outstanding token work in each queue.

Our technique is practical and delivers high performance.
We implemented our prototype system (called \system) on top of vLLM version 0.92,
without modifying memory management or model specification.
Evaluation on four NVIDIA A100 GPUs with state-of-the-art open-weight models shows that \system delivers 1.28$\times$--4.52$\times$ higher throughput than GPU-level splits. Chunked and single-pass embeddings are equivalent up to floating-point reduction-order effects (\cref{tab:pooling-equivalence}).
%

This work focuses on compute patterns inside GPU kernels, orthogonal to parameter sharing~\citep{vasani2025phlora, gauthier2024lord} and multi-adapter scheduling~\citep{sheng_slora_2024, chen_punica_2024}.

\section{Background}
\label{sec:background}

\subsection{Modern RAG Workflow}
\label{sec:background:rag}

Modern RAG pipelines combine embedding-based retrieval with language model generation~\citep{lewis_retrieval-augmented_2020}.
Recent techniques employ iterative retrieval, where embedding and generation alternate to refine queries dynamically~\citep{asai_self-rag_2023, jiang_active_2023, shao_enhancing_2023, trivedi_interleaving_2023}.
A single prompt thus decomposes into a sequence of embedding and generation requests, creating a mixed workload.
\blue{\system changes model-side embedding inference but not the vector-search
	backend, which is orthogonal to our serving design.}

\subsection{Computational Characteristics of Embedding and Generation}
\label{sec:background:characteristics}

Embedding and generation requests differ fundamentally in their execution patterns and resource utilization.

\paragraph{Generation}
A generation request produces a variable-length output autoregressively.
It executes in two phases: \emph{prefill} and \emph{decode}.
The \emph{prefill} phase processes the input prompt in a single parallel \emph{forward} pass.
The \emph{decode} phase then generates tokens one at a time: each step reads the key-value (KV) cache~\citep{pope_efficiently_2022} containing previous tokens' representations, computes \emph{attention}, and appends the new token's KV entries.
This decode phase is \emph{memory-bound}, as each step transfers large KV cache while performing relatively few arithmetic operations~\citep{sheng_flexgen_2023}.

\paragraph{Embedding}
An embedding request structurally resembles the \emph{prefill} phase of generation, but follows with a pooling operation rather than entering a decode loop.
The model processes input tokens through transformer layers~\citep{vaswani_attention_2017} and aggregates the final hidden states via pooling (typically mean pooling or last-token pooling) to produce a fixed-length vector~\citep{wang_improving_2024, shen_retrieval-augmented_2024}.
A vector database indexes these vectors for similarity search~\citep{hnsw_2020, faiss, milvus_2021, airindex, cloudindex}.
Because embedding execution is a single dense forward pass, it is \emph{compute-bound}~\citep{ivanov_data_2021}, contrasting with memory-bound decoding.

\begin{table}[t]
	\scriptsize
	\centering
	\caption{GPU utilization under pure embedding and pure generation workloads (Mistral-7B on A100-40GB, 128-token inputs, 512-token outputs for generation).}
	\vspace{-1mm}
	\label{tab:gpu-util}
	\begin{tabular}{l|cc|cc}
		\toprule
		           & \multicolumn{2}{c|}{GPU Compute (\%)} & \multicolumn{2}{c}{GPU Memory (\%)}                        \\
		Workload   & Mean                                  & Max                                 & Mean          & Max  \\
		\midrule
		Embedding  & \textbf{85.3}                         & 98.0                                & 36.5          & 37.1 \\
		Generation & 75.5                                  & 98.0                                & \textbf{90.5} & 90.5 \\
		\bottomrule
	\end{tabular}
	\vspace{-4mm}
\end{table}

\paragraph{Resource utilization analysis}
To quantify these compute differences, we profiled GPU utilization under pure embedding and pure generation workloads using Mistral-7B on an A100-40GB GPU.
\cref{tab:gpu-util} summarizes the results.
Neither workload fully exploits both resources, suggesting an opportunity to co-schedule them for improved efficiency.

\paragraph{Parameter sharing}
State-of-the-art embedding models are increasingly derived from generative models via LoRA fine-tuning~\citep{li_towards_2023, wang_improving_2024, yavuz_sfr-embedding-mistral_2024, ma_fine-tuning_2024, behnamghader_llm2vec_2024}.
Other approaches train unified models for both tasks~\citep{muennighoff_generative_2024, zhang_gem_2025, li_unigen_2024, tang_boosting_2025}.
\blue{\system is optimized for shared-backbone embedding and generation
	models. For separate backbones, our extension retains \ibs across
	model-specific queues but does not combine the models in one forward pass
	(\cref{tab:separate-model}).}

\subsection{Batching and Scheduling}
\label{sec:background:batching}

Modern LLM serving systems employ \emph{iteration-level scheduling}~\citep{yu2022orca, kwon_efficient_2023}, re-evaluating batch composition at every token generation step.
Within each iteration, the system executes a \emph{homogeneous batch} consisting exclusively of generation requests, potentially mixing \blue{chunked} prefill and decode phases~\citep{agrawal_taming_2024}.
\blue{\system extends this chunked execution to embeddings via incremental pooling.}
Recent work targets scheduling policies~\citep{wu_fast_2024, sheng_fairness_2024}, kernel efficiency \blue{and GPU co-location}~\citep{wang_lightseq_2021, xia2023flashllm, orion_2024}, KV cache management~\citep{mooncake_2025, pensieve_2025, hcache_2024, cacheCraft_2025, xia_lazyattention_2026}, storage efficiency~\cite{qstore}, and prefill--decode disaggregation~\citep{zhong_distserve_2024, hu_inference_2024}.
Multi-tenant LoRA systems~\citep{sheng_slora_2024, chen_punica_2024} improve adapter throughput but assume generation-only workloads. \blue{\system schedules embedding and generation jointly while preserving arrival order within each request type (\cref{sec:scheduling}).}

\section{Unified Runner}
\label{sec:batch}


This section presents the design of \system's unified runner, which enables heterogeneous batching of embedding and generation requests within a single inference loop.


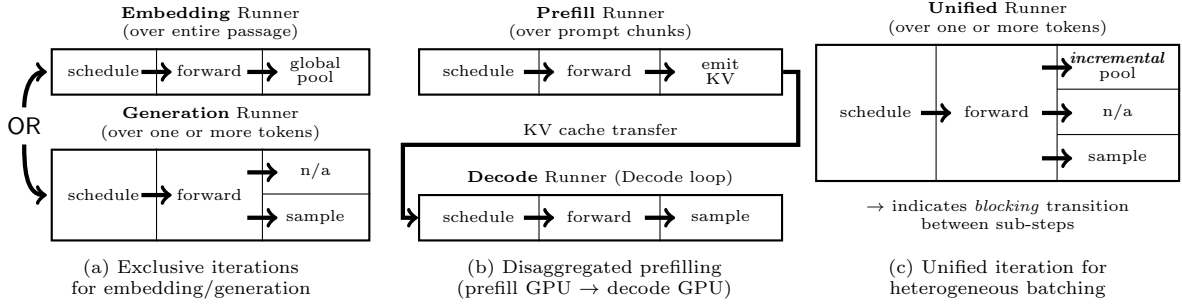
\begin{figure*}[t]

	\tikzset{
		itr/.style={
				draw=black, thick, minimum width=3*\cellwidth, minimum height=\cellheight
			}
	}

	\centering
	\captionsetup[subfigure]{justification=centering,font=scriptsize}
	\captionsetup{font=small}
	\hfill
	\begin{subfigure}[b]{0.32\linewidth}
		\centering
		\begin{tikzpicture}

			\def\cellwidth{14mm}
			\def\cellheight{6mm}
			\def\arrowlen{4mm}

			\node[itr] (E) {};
			\node[font=\tiny,inner ysep=0,anchor=south,align=center,text width=48mm]
			at ($(E.north)+(0,0.1)$) {\textbf{Embedding} Runner \\ (over entire passage)};
			\draw[-] ($(E.north west)+(\cellwidth,0)$) -- ($(E.south west)+(\cellwidth,0)$);
			\draw[-] ($(E.north west)+(2*\cellwidth,0)$) -- ($(E.south west)+(2*\cellwidth,0)$);
			\node[font=\tiny] at ($(E.north west)+(0.5*\cellwidth,-0.5*\cellheight)$) {schedule};
			\node[font=\tiny] at ($(E.north west)+(1.5*\cellwidth,-0.5*\cellheight)$) {forward};
			\node[font=\tiny,align=center] at ($(E.north west)+(2.5*\cellwidth,-0.5*\cellheight)$)
			{global\\[-0.2em] pool};

			\draw[->,ultra thick] ($(E.west)+(\cellwidth - 0.5*\arrowlen,0)$) -- ++(\arrowlen,0);
			\draw[->,ultra thick] ($(E.west)+(2*\cellwidth - 0.5*\arrowlen,0)$) -- ++(\arrowlen,0);

			\node[itr,minimum height=2*\cellheight,anchor=north] (G) at ($(E.south)+(0,-0.7)$) {};
			\node[font=\tiny,inner ysep=0,anchor=south,align=center,text width=48mm]
			at ($(G.north)+(0,0.1)$) {\textbf{Generation} Runner \\ (over one or more tokens)};
			\draw[-] ($(G.north west)+(\cellwidth,0)$) -- ($(G.south west)+(\cellwidth,0)$);
			\draw[-] ($(G.north west)+(2*\cellwidth,0)$) -- ($(G.south west)+(2*\cellwidth,0)$);
			\draw[-] ($(G.north west)+(2*\cellwidth,-\cellheight)$) -- ($(G.north east)+(0,-\cellheight)$);
			\node[font=\tiny] at ($(G.north west)+(0.5*\cellwidth,-1*\cellheight)$) {schedule};
			\node[font=\tiny] at ($(G.north west)+(1.5*\cellwidth,-1*\cellheight)$) {forward};
			\node[font=\tiny] at ($(G.north west)+(2.5*\cellwidth,-0.5*\cellheight)$) {n/a};
			\node[font=\tiny] at ($(G.north west)+(2.5*\cellwidth,-1.5*\cellheight)$) {sample};

			\draw[->,ultra thick] ($(G.west)+(\cellwidth - 0.5*\arrowlen,0)$) -- ++(\arrowlen,0);
			\draw[->,ultra thick] ($(G.west)+(0,0.5*\cellheight)+(2*\cellwidth - 0.5*\arrowlen,0)$) -- ++(\arrowlen,0);
			\draw[->,ultra thick] ($(G.west)+(0,-0.5*\cellheight)+(2*\cellwidth - 0.5*\arrowlen,0)$) -- ++(\arrowlen,0);

			\node[font=\footnotesize\sffamily] at ($(G.north west)+(-0.35,+0.3)$) (or) {OR};
			\draw[ultra thick,->] (or.north) to[out=90,in=180] ($(E.west)+(-0.05,0)$);
			\draw[ultra thick,->] (or.south) to[out=270,in=180] ($(G.west)+(-0.05,0)$);

		\end{tikzpicture}

		\caption{Exclusive iterations \\ for embedding/generation}
	\end{subfigure}
	\hfill
	\begin{subfigure}[b]{0.32\linewidth}
		\centering
		\begin{tikzpicture}

			\def\cellwidth{16mm}
			\def\cellheight{6mm}
			\def\arrowlen{4mm}

			\node[itr] (P) {};
			\node[font=\tiny,inner ysep=0,anchor=south,align=center,text width=48mm]
			at ($(P.north)+(0,0.1)$) {\textbf{Prefill} Runner \\ (over prompt chunks)};

			\draw[-] ($(P.north west)+(\cellwidth,0)$) -- ($(P.south west)+(\cellwidth,0)$);
			\draw[-] ($(P.north west)+(2*\cellwidth,0)$) -- ($(P.south west)+(2*\cellwidth,0)$);

			\node[font=\tiny] at ($(P.north west)+(0.5*\cellwidth,-0.5*\cellheight)$) {schedule};
			\node[font=\tiny] at ($(P.north west)+(1.5*\cellwidth,-0.5*\cellheight)$) {forward};
			\node[font=\tiny,align=center] at ($(P.north west)+(2.5*\cellwidth,-0.5*\cellheight)$)
			{emit\\[-0.2em] KV};

			\draw[->,ultra thick] ($(P.west)+(\cellwidth - 0.5*\arrowlen,0)$) -- ++(\arrowlen,0);
			\draw[->,ultra thick] ($(P.west)+(2*\cellwidth - 0.5*\arrowlen,0)$) -- ++(\arrowlen,0);

			\node[itr,anchor=north] (D) at ($(P.south)+(0,-1.3)$) {};
			\node[font=\tiny,inner ysep=0,anchor=south,align=center,text width=48mm]
			at ($(D.north)+(0,0.1)$) {\textbf{Decode} Runner (Decode loop)};

			\draw[-] ($(D.north west)+(\cellwidth,0)$) -- ($(D.south west)+(\cellwidth,0)$);
			\draw[-] ($(D.north west)+(2*\cellwidth,0)$) -- ($(D.south west)+(2*\cellwidth,0)$);

			\node[font=\tiny] at ($(D.north west)+(0.5*\cellwidth,-0.5*\cellheight)$) {schedule};
			\node[font=\tiny] at ($(D.north west)+(1.5*\cellwidth,-0.5*\cellheight)$) {forward};
			\node[font=\tiny] at ($(D.north west)+(2.5*\cellwidth,-0.5*\cellheight)$) {sample};

			\draw[->,ultra thick] ($(D.west)+(\cellwidth - 0.5*\arrowlen,0)$) -- ++(\arrowlen,0);
			\draw[->,ultra thick] ($(D.west)+(2*\cellwidth - 0.5*\arrowlen,0)$) -- ++(\arrowlen,0);

			\draw[ultra thick, black!70!black, ->]
			(P.east) -- ++(0.2,0)          
			|- ($(P.south)!0.5!(D.north)$) 
			-| ($(D.west)+(-0.2,0)$)       
			-- (D.west);                   

			\node[font=\tiny, anchor=south, align=center] at ($(P.south)+(0.0, -0.65)$)
			{KV cache transfer};

		\end{tikzpicture}
		\caption{Disaggregated prefilling \\ (prefill GPU $\rightarrow$ decode GPU)}
	\end{subfigure}
	\hfill
	\begin{subfigure}[b]{0.32\linewidth}
		\centering
		\begin{tikzpicture}

			\def\cellwidth{16mm}
			\def\cellheight{6mm}
			\def\arrowlen{4mm}

			\node[itr,minimum height=3*\cellheight] (E) {};
			\node[font=\tiny,inner ysep=0,anchor=south,align=center,text width=48mm]
			at ($(E.north)+(0,0.1)$) {\textbf{Unified} Runner \\ (over one or more tokens)};
			\draw[-] ($(E.north west)+(\cellwidth,0)$) -- ($(E.south west)+(\cellwidth,0)$);
			\draw[-] ($(E.north west)+(2*\cellwidth,0)$) -- ($(E.south west)+(2*\cellwidth,0)$);
			\draw[-] ($(E.north west)+(2*\cellwidth,-\cellheight)$) -- ($(E.north east)+(0,-\cellheight)$);
			\draw[-] ($(E.north west)+(2*\cellwidth,-2*\cellheight)$) -- ($(E.north east)+(0,-2*\cellheight)$);
			\node[font=\tiny] at ($(E.north west)+(0.5*\cellwidth,-1.5*\cellheight)$) {schedule};
			\node[font=\tiny] at ($(E.north west)+(1.5*\cellwidth,-1.5*\cellheight)$) {forward};
			\node[font=\tiny,align=center] at ($(E.north west)+(2.5*\cellwidth,-0.5*\cellheight)$)
			{\textbf{\emph{incremental}}\\[-0.2em] pool};
			\node[font=\tiny] at ($(E.north west)+(2.5*\cellwidth,-1.5*\cellheight)$) {n/a};
			\node[font=\tiny] at ($(E.north west)+(2.5*\cellwidth,-2.5*\cellheight)$) {sample};

			\draw[->,ultra thick] ($(E.west)+(\cellwidth - 0.5*\arrowlen,0)$) -- ++(\arrowlen,0);
			\draw[->,ultra thick] ($(E.west)+(2*\cellwidth - 0.5*\arrowlen,-1*\cellheight)$) -- ++(\arrowlen,0);
			\draw[->,ultra thick] ($(E.west)+(2*\cellwidth - 0.5*\arrowlen,0)$) -- ++(\arrowlen,0);
			\draw[->,ultra thick] ($(E.west)+(2*\cellwidth - 0.5*\arrowlen,1*\cellheight)$) -- ++(\arrowlen,0);

			\node[font=\tiny,anchor=south,align=center,text width=48mm,draw=none]
			(L) at ($(E.south)+(0,-0.8)$)
			{\textbf{$\rightarrow$} indicates \emph{blocking} transition \\ between sub-steps};

		\end{tikzpicture}
		\caption{Unified iteration for \\ heterogeneous batching}
	\end{subfigure}
	\hfill

	\caption{Runner structure comparison. (a) Task-separated runners use global
		pooling for embedding or token sampling for generation. (b) Phase-separated
		prefill and decode runners transfer the KV cache between GPUs. (c) Our
		unified runner supports both output paths in each iteration; incremental
		pooling allows embedding chunks to execute alongside generation requests.}
	\label{fig:runner}
\end{figure*}

\subsection{High-Level Workflow}
\label{sec:batch:workflow}

Our runner is designed for iteration-level
scheduling~\citep{yu2022orca}, aiming for seamless integration with
vLLM's~\citep{kwon_efficient_2023} open-sourced model management and server.
Yet, our approach differs from existing approaches in that both
embedding and generation requests are processed using the same
runner, which we achieve by consolidating distinct embedding and
generation runners into one in a structured way.

Our runner operates in three blocking stages: (1) \texttt{schedule},
(2) \texttt{forward}, and (3) \texttt{emit} (\cref{fig:runner}).
\texttt{Schedule} determines batch composition, including generation prefills, decodes, or embedding.
\texttt{Forward} applies attention and feed-forward layers, identical across request types except for input length.
\texttt{Emit} diverges: generation samples the next token, while embedding performs \emph{incremental pooling} to compute the final embedding vector.



\begin{algorithm}[t]
	\scriptsize
	\caption{Heterogeneous Batching Inference Loop}
	\label{alg:hetero-batch}
	\KwIn{Embedding queue $Q_e$, generation queue $Q_g$}
	\While{$Q_e \neq \emptyset$ \textbf{or} $Q_g \neq \emptyset$ \textbf{or} in-flight requests}{
		$\mathcal{B} \gets \text{\textsc{Schedule}}(Q_e, Q_g)$ \tcp*{Mix embed \& gen}
		$H \gets \text{\textsc{Forward}}(\mathcal{B})$ \tcp*{Attention + FFN}
		\tcp{Parallel output processing}
		\Begin(in parallel){
			$\mathcal{R}_e \gets \text{\textsc{IncPooler}}(H[\text{emb\_idx}])$\;
			$\mathcal{R}_g \gets \text{\textsc{Sampler}}(H[\text{gen\_idx}])$\;
		}
		$\text{\textsc{Emit}}(\mathcal{R}_e, \mathcal{R}_g)$ \tcp*{Return completed}
	}
\end{algorithm}
\vspace{-1mm}

\subsection{Batch Unit}
\label{sec:batch:unit}

A batch in \system contains three types of items: (1) \emph{embedding chunks}, which are segments of an embedding request's input sequence, (2) \emph{generation prefill chunks}, which are segments of a generation request's prompt, and (3) \emph{generation decode tokens}, which are single tokens generated during autoregressive decoding. We call execution of an embedding request across multiple such chunks \emph{chunked embedding}.
All three types share the same forward pass through the transformer layers, differing only in their input lengths and output handling.
Embedding and prefill chunks process multiple tokens per request, while decode tokens process exactly one token per request.
This unified representation allows the scheduler to mix request types freely within a single batch, filling available token capacity while preserving the workload-proportional allocation described in \cref{sec:scheduling:design}.
The scheduler outputs batches sorted by request type, ensuring that items of the same type are contiguous in memory.
This layout eliminates memory movement when invoking the output kernels: the incremental pooler, the sampler, and the prefill handler each operate on a contiguous slice of the hidden state tensor, avoiding gather operations or intermediate copies.

\subsection{Incremental Pooler}
\label{sec:batch:pooler}

Incremental pooling is the cross-chunk aggregation used by chunked embedding.


Consider an embedding request with input sequence of $n$ tokens.
Under chunked embedding, the system splits the sequence into $K$ chunks of sizes $c_1, c_2, \ldots, c_K$ where $\sum_{k=1}^{K} c_k = n$.
At iteration $k$, the forward pass produces hidden states $H^{(k)} \in \mathbb{R}^{c_k \times d}$ for the $k$-th chunk, where $d$ is the hidden dimension.

\blue{\paragraph{Pooling Heads}
Mean and weighted-mean pooling share the same streaming formulation. For
weights $w_{k,j}$, the pooler maintains

{\small
\begin{align}
	S^{(k)} &= S^{(k-1)} + \sum_{j=1}^{c_k} w_{k,j}H_j^{(k)}, & S^{(0)} &= \mathbf{0}, \\
	W^{(k)} &= W^{(k-1)} + \sum_{j=1}^{c_k} w_{k,j}, & W^{(0)} &= 0,
\end{align}
}%
and emits $S^{(K)}/W^{(K)}$ after the final chunk. 
Mean pooling is the special case $w_{k,j}=1$. 
Positional heads retain only the selected hidden state:
CLS pooling uses $H_1^{(1)}$, while last-token pooling uses
$H_{c_K}^{(K)}$. 
These additive and positional heads require $O(d)$ state per
request. An arbitrary deterministic head remains compatible by buffering
$A^{(k)}=[A^{(k-1)};H^{(k)}]$ and applying the head after the final chunk, at
an $O(nd)$ memory cost.}
\blue{For all of these heads, incremental and single-pass pooling are
algebraically equivalent, although floating-point non-associativity may
introduce small numerical differences.}

\section{\ibsfull}
\label{sec:scheduling}

We present \ibsfull (\ibs), a scheduling algorithm that dynamically allocates capacity proportional to the workload mix.
We first define three requirements for efficient scheduling (\cref{sec:scheduling:requirements}), then detail the \ibs design (\cref{sec:scheduling:design}).

\subsection{Requirements}
\label{sec:scheduling:requirements}

We define three requirements:
(1) \emph{Proportional allocation}---capacity should match the instantaneous embedding-to-generation demand ratio;
(2) \emph{Per-type FCFS}---requests of the same type are served in arrival order, though types may interleave;
(3) \emph{No head-of-line blocking}---neither request type should starve the other.

%
%

\subsection{Scheduler Design}
\label{sec:scheduling:design}

\begin{algorithm}[t]
	\scriptsize
	\caption{\ibsfull}
	\label{alg:ibs}
	\KwIn{Embedding queue $Q_e$, generation queue $Q_g$, token budget $B$,
		\blue{persistent credits $a_e,a_g$}}
	\KwOut{Scheduled batch $\mathcal{B}$}
	$(\mathcal{B}, u) \gets \text{\textsc{ScheduleRunning}}(B)$\;
	$B \gets B-u$\;
	\tcp{Proportionally select from both queues}
	$w_e \gets \sum_{r \in Q_e}\mathrm{tokens}(r)$; \quad
		$w_g \gets \sum_{r \in Q_g}\mathrm{tokens}(r)$\;
	\If{$Q_e = \emptyset$}{$a_e \gets 0$}
		\If{$Q_g = \emptyset$}{$a_g \gets 0$}
	\While{$B > 0$ \textbf{and} $(Q_e \neq \emptyset$ \textbf{or} $Q_g \neq \emptyset)$}{
		$a_e \gets a_e + w_e$; \quad $a_g \gets a_g + w_g$\;
		\uIf{$Q_g = \emptyset$ \textbf{or} $(Q_e \neq \emptyset$ \textbf{and} $a_e \geq a_g)$}{
			$(\mathcal{I}, u) \gets \text{\textsc{ScheduleOne}}(Q_e, B)$\;
			$\mathcal{B} \gets \mathcal{B} \cup \mathcal{I}$; $B \gets B-u$\;
			$a_e \gets a_e - (w_e + w_g)$\;
		}
		\Else{
			$(\mathcal{I}, u) \gets \text{\textsc{ScheduleOne}}(Q_g, B)$\;
			$\mathcal{B} \gets \mathcal{B} \cup \mathcal{I}$; $B \gets B-u$\;
			$a_g \gets a_g - (w_e + w_g)$\;
		}
	}
	\tcp{$a_e, a_g$ persist across iterations; reset above when a queue drains}
	\Return{$\mathcal{B}$}
\end{algorithm}

\cref{alg:ibs} presents our intra-batch scheduling algorithm, which satisfies all three requirements.
The scheduler first reserves capacity for in-flight generation requests that are in the decode phase.
These requests have allocated KV cache entries and must continue execution to avoid KV cache movement.
This phase mirrors vLLM scheduling and uses the same swap-in and swap-out mechanisms.

The scheduler then fills the remaining budget by alternating between the embedding and generation queues.
\blue{\paragraph{Why chunking improves packing.}
Let $R$ be the residual token budget after reserving capacity for in-flight
decode requests. Without chunking, an embedding request of length $e$ is
admitted only when $e \leq R$; otherwise, the remaining $R$ tokens go unused.
\textsc{ScheduleOne} instead admits embedding chunks of at most $C$ tokens.
When token capacity is the limiting constraint, a backlogged embedding queue
therefore leaves fewer than $C$ tokens unused once no additional chunk fits.}
\blue{Let $w_e$ and $w_g$ be the sums of the remaining token counts in the
embedding and generation queues, respectively.}
The algorithm maintains accumulators $a_e$ and $a_g$ that track the scheduling credit of each queue.
\blue{The credits persist across scheduling iterations while their queues remain nonempty.}
At each selection step, the scheduler increments both accumulators by their corresponding weights and selects a request from the queue with the larger accumulator, decrementing the selected accumulator by $w_e + w_g$.
If one queue becomes empty, the scheduler continues selecting from the remaining queue until the token budget is exhausted.

The accumulator mechanism ensures that the ratio of scheduled embeddings to generations converges to $w_e : w_g$, achieving proportional allocation that matches the instantaneous workload mix. When only embeddings are queued ($w_g = 0$), all capacity goes to embeddings, and vice versa.
Finally, by maintaining separate queues and alternating between them proportionally, the scheduler avoids head-of-line blocking. Short embedding requests are never stuck behind long generation requests, and each type makes progress according to its share of the workload.
\blue{Assume each nonempty queue has positive weight, its head request or next
chunk can eventually be admitted, and each iteration completes in bounded
time. Persistent credits then prevent scheduler starvation: a backlogged queue
that is not selected continues accumulating credit until it is chosen. For
fixed weights, its allocation differs from its ideal proportional share by at
most one scheduling opportunity. This guarantee addresses scheduler-induced
blocking. It does not bound queuing delay when offered load exceeds capacity.}

\section{Evaluation}
\label{sec:eval}






We evaluate \system, an embedding-generation serving system that combines heterogeneous batching with \ibsfull (\ibs).
The setup is described in \cref{sec:eval:setup}.
Our results show that:

\begin{itemize}[left=1mm,topsep=0mm,itemsep=0mm]
	\item \textbf{\emph{High Throughput:}} 1.28$\times$--4.52$\times$ higher throughput than GPU-level splits on synthetic workloads; 2.4$\times$ on a RAG benchmark with 4 GPUs (\cref{sec:eval:throughput}).
	\item \textbf{\emph{Low Latency:}} 9\% lower p99 generation and 16\% lower p99 embedding latency; up to 55.8\% lower p99 on RAG (\cref{sec:eval:starvation}).
	\item \textbf{\emph{High GPU Utilization:}} 41 pp higher average utilization, completing mixed workloads 43\% faster (\cref{sec:eval:load_balance}).
	\item \textbf{\emph{Negligible Overhead:}} At saturation, matches dedicated-engine throughput for both workload types (\cref{sec:eval:tax}).
	\item \textbf{\emph{Generalizability:}} Validated on 3 models with LoRA ranks up to 64 (\cref{sec:eval:generalizability}).
\end{itemize}

\begin{figure*}[ht]
	\captionsetup[subfigure]{justification=centering}
	\centering
	\begin{subfigure}[b]{0.32\linewidth}
		\centering
\begin{tikzpicture}
  \begin{axis}[
      width=\linewidth,
      height=4.1cm,
      label style={font=\scriptsize, align=center},
      tick label style={font=\scriptsize},
      xlabel={Ratio of Generation Requests},
      ylabel={Throughput (req/s)},
      ymin=0,
      ymax=100,
      xtick={0.0, 0.5, 1.0},
      legend style={
        legend cell align={left},
        font=\scriptsize,
        legend columns=4,
        column sep=0.8em,
        at={(0, 1.1)},
        anchor=south west,
      },
      mark options={scale=0.6},
    ]

    \addplot+[mark=o, color=blue]   table[col sep=comma, x=ratio_gen,
    y=baseline_2_2_combined] {figures/data/throughput_vs_ratio_orig.csv};
    \addlegendentry{$\text{\dedicated}_{\text{2:2}}$}

    \addplot+[mark=o, color=navy]   table[col sep=comma, x=ratio_gen,
    y=baseline_1_3_combined] {figures/data/throughput_vs_ratio_orig.csv};
    \addlegendentry{$\text{\dedicated}_{\text{1:3}}$}

    \addplot+[mark=o, color=purple] table[col sep=comma, x=ratio_gen,
    y=baseline_3_1_combined] {figures/data/throughput_vs_ratio_orig.csv};
    \addlegendentry{$\text{\dedicated}_{\text{3:1}}$}

    \addlegendimage{area legend, draw=none, fill=blue, fill opacity=0.40}
    \addlegendentry{Generation Requests}

    \addplot+[mark=o, color=orange] table[col sep=comma, x=ratio_gen,
    y=hybrid_turn_combined] {figures/data/throughput_vs_ratio_orig.csv};
    \addlegendentry{\sharedhomo}

    \addplot+[mark=o, color=yellow] table[col sep=comma, x=ratio_gen,
    y=hybrid_fcfs_combined] {figures/data/throughput_vs_ratio_orig.csv};
    \addlegendentry{\sharedhetero}

    \addplot+[mark=o, color=red, solid, thick]  table[col sep=comma,
    x=ratio_gen, y=hybrid_split_combined] {figures/data/throughput_vs_ratio_orig.csv};
    \addlegendentry{\textbf{\system}}

    \addlegendimage{area legend, draw=none, fill=red, fill opacity=0.60}
    \addlegendentry{Embedding Requests}

  \end{axis}
\end{tikzpicture}
		\vspace{-30pt}
		\caption{Overall Throughput\\~}
		\label{fig:throughput_ratio}
	\end{subfigure}
	\hfill
	\begin{subfigure}[b]{0.32\linewidth}
		\centering
		\begin{tikzpicture}
  \begin{groupplot}[
      group style={
        group size=1 by 4,
        vertical sep=7pt,
        x descriptions at=edge bottom,
      },
      width=\linewidth,
      height=2.0cm,
      xmin=0, xmax=120,
      ymin=0, ymax=200,
      xlabel={Time (s)},
      xtick={0, 60, 120},
      grid=both,
      label style={font=\scriptsize},
      tick label style={font=\scriptsize},
      legend style={
        legend cell align={left},
        font=\scriptsize,
        legend columns=2,
        column sep=0.5em,
        at={(-1, 1.1)},
        anchor=south west,
      },
    ]
    \pgfplotsset{every axis title/.append style={at={(1,-0.7)},
    font=\scriptsize, anchor=south east}}

    \nextgroupplot[
      ytick={0, 200},
      title=GPU0,
      title style={at={(1,-0.4)}, font=\scriptsize, anchor=south east},
    ]
    \addplot[name path=zero, draw=none] {0};
    \addplot[name path=E0, const plot, draw=none]
    table[col sep=comma, x=t,
    y=active_embed]{figures/data/baseline_timeline_gpu_0.csv};
    \addplot[name path=C0, const plot, draw=none]
    table[col sep=comma, x=t, y
    expr=\thisrow{active_embed}+\thisrow{active_gen}]
    {figures/data/baseline_timeline_gpu_0.csv};
    \addplot[fill=red,  fill opacity=0.40, draw=none] fill
    between[of=E0 and zero];
    \addplot[fill=blue,   fill opacity=0.40, draw=none] fill
    between[of=C0 and E0];

    \nextgroupplot[
      ytick={0, 200},
      title=GPU1,
    ]
    \addplot[name path=zero, draw=none] {0};
    \addplot[name path=E1, const plot, draw=none]
    table[col sep=comma, x=t,
    y=active_embed]{figures/data/baseline_timeline_gpu_1.csv};
    \addplot[name path=C1, const plot, draw=none]
    table[col sep=comma, x=t, y
    expr=\thisrow{active_embed}+\thisrow{active_gen}]
    {figures/data/baseline_timeline_gpu_1.csv};
    \addplot[fill=red,  fill opacity=0.40, draw=none] fill
    between[of=E1 and zero];
    \addplot[fill=blue,   fill opacity=0.40, draw=none] fill
    between[of=C1 and E1];

    \nextgroupplot[ylabel={\# of Active requests},
      every axis y label/.append style={at=(ticklabel cs:1.1)},
      ytick={0, 200},
      title=GPU2,
    ]
    \addplot[name path=zero, draw=none] {0};
    \addplot[name path=E2, const plot, draw=none]
    table[col sep=comma, x=t,
    y=active_embed]{figures/data/baseline_timeline_gpu_2.csv};
    \addplot[name path=C2, const plot, draw=none]
    table[col sep=comma, x=t, y
    expr=\thisrow{active_embed}+\thisrow{active_gen}]
    {figures/data/baseline_timeline_gpu_2.csv};
    \addplot[fill=red,  fill opacity=0.40, draw=none] fill
    between[of=E2 and zero];
    \addplot[fill=blue,   fill opacity=0.40, draw=none] fill
    between[of=C2 and E2];

    \nextgroupplot[xlabel={Time (s)}, ytick={0, 200}, title=GPU3]
    \addplot[name path=zero, draw=none] {0};
    \addplot[name path=E3, const plot, draw=none]
    table[col sep=comma, x=t,
    y=active_embed]{figures/data/baseline_timeline_gpu_3.csv};
    \addplot[name path=C3, const plot, draw=none]
    table[col sep=comma, x=t, y
    expr=\thisrow{active_embed}+\thisrow{active_gen}]
    {figures/data/baseline_timeline_gpu_3.csv};
    \addplot[fill=red,  fill opacity=0.40, draw=none] fill
    between[of=E3 and zero];
    \addplot[fill=blue,   fill opacity=0.40, draw=none] fill
    between[of=C3 and E3];

  \end{groupplot}
\end{tikzpicture}
		\vspace{-30pt}
		\caption{Active Requests over time\\($\text{\dedicated}_{\text{1:3}}$)}
		\label{fig:baseline_timeline}
	\end{subfigure}
	\hfill
	\begin{subfigure}[b]{0.32\linewidth}
		\centering
		\begin{tikzpicture}
  \begin{groupplot}[
      group style={
        group size=1 by 4,
        vertical sep=7pt,
        x descriptions at=edge bottom,
      },
      width=\linewidth,
      height=2.0cm,
      xmin=0, xmax=120,
      ymin=0, ymax=200,
      xlabel={Time (s)},
      xtick={0, 60, 120},
      grid=both,
      label style={font=\scriptsize},
      tick label style={font=\scriptsize},
      legend style={
        legend cell align={left},
        font=\scriptsize,
        legend columns=1,
        column sep=0.5em,
        at={(-0.05, 1.45)},
        anchor=south west,
      },
    ]
    \pgfplotsset{every axis title/.append style={at={(1,-0.7)},
    font=\scriptsize, anchor=south east}}

    \nextgroupplot[ytick={0, 200}, title=GPU0]
    \addplot[name path=G0lo, const plot, draw=none]
    table[col sep=comma, x=t, y=gen_y0]{figures/data/hybrid_timeline_gpu_0.csv};
    \addplot[name path=G0hi, const plot, draw=none]
    table[col sep=comma, x=t, y=gen_y1]{figures/data/hybrid_timeline_gpu_0.csv};

    \addplot[name path=E0lo, const plot, draw=none]
    table[col sep=comma, x=t,
    y=embed_y0]{figures/data/hybrid_timeline_gpu_0.csv};
    \addplot[name path=E0hi, const plot, draw=none]
    table[col sep=comma, x=t,
    y=embed_y1]{figures/data/hybrid_timeline_gpu_0.csv};

    \addplot[fill=blue, fill opacity=0.40, draw=none]  fill
    between[of=G0hi and G0lo];
    \addplot[fill=red,  fill opacity=0.60, draw=none]  fill
    between[of=E0hi and E0lo];

    \nextgroupplot[ytick={0, 200}, title=GPU1]
    \addplot[name path=G1lo, const plot, draw=none]
    table[col sep=comma, x=t, y=gen_y0]{figures/data/hybrid_timeline_gpu_1.csv};
    \addplot[name path=G1hi, const plot, draw=none]
    table[col sep=comma, x=t, y=gen_y1]{figures/data/hybrid_timeline_gpu_1.csv};
    \addplot[name path=E1lo, const plot, draw=none]
    table[col sep=comma, x=t,
    y=embed_y0]{figures/data/hybrid_timeline_gpu_1.csv};
    \addplot[name path=E1hi, const plot, draw=none]
    table[col sep=comma, x=t,
    y=embed_y1]{figures/data/hybrid_timeline_gpu_1.csv};
    \addplot[fill=blue, fill opacity=0.40, draw=none]  fill
    between[of=G1hi and G1lo];
    \addplot[fill=red,  fill opacity=0.60, draw=none]  fill
    between[of=E1hi and E1lo];

    \nextgroupplot[ylabel={\# of Active requests},
      every axis y label/.append style={at=(ticklabel cs:1.1)},
    ytick={0, 200}, title=GPU2]
    \addplot[name path=G2lo, const plot, draw=none]
    table[col sep=comma, x=t, y=gen_y0]{figures/data/hybrid_timeline_gpu_2.csv};
    \addplot[name path=G2hi, const plot, draw=none]
    table[col sep=comma, x=t, y=gen_y1]{figures/data/hybrid_timeline_gpu_2.csv};
    \addplot[name path=E2lo, const plot, draw=none]
    table[col sep=comma, x=t,
    y=embed_y0]{figures/data/hybrid_timeline_gpu_2.csv};
    \addplot[name path=E2hi, const plot, draw=none]
    table[col sep=comma, x=t,
    y=embed_y1]{figures/data/hybrid_timeline_gpu_2.csv};
    \addplot[fill=blue, fill opacity=0.40, draw=none]  fill
    between[of=G2hi and G2lo];
    \addplot[fill=red,  fill opacity=0.60, draw=none]  fill
    between[of=E2hi and E2lo];

    \nextgroupplot[xlabel={Time (s)}, ytick={0, 200}, title=GPU3]
    \addplot[name path=G3lo, const plot, draw=none]
    table[col sep=comma, x=t, y=gen_y0]{figures/data/hybrid_timeline_gpu_3.csv};
    \addplot[name path=G3hi, const plot, draw=none]
    table[col sep=comma, x=t, y=gen_y1]{figures/data/hybrid_timeline_gpu_3.csv};
    \addplot[name path=E3lo, const plot, draw=none]
    table[col sep=comma, x=t,
    y=embed_y0]{figures/data/hybrid_timeline_gpu_3.csv};
    \addplot[name path=E3hi, const plot, draw=none]
    table[col sep=comma, x=t,
    y=embed_y1]{figures/data/hybrid_timeline_gpu_3.csv};
    \addplot[fill=blue, fill opacity=0.40, draw=none]  fill
    between[of=G3hi and G3lo];
    \addplot[fill=red,  fill opacity=0.60, draw=none]  fill
    between[of=E3hi and E3lo];

  \end{groupplot}
\end{tikzpicture}
		\vspace{-30pt}
		\caption{Active Requests over time\\(\system)}
		\label{fig:hybrid_timeline}
	\end{subfigure}

	\caption{Throughput and GPU utilization comparison.
	(\cref{fig:throughput_ratio}) compares throughput of existing serving techniques vs our heterogeneous batching, \system (see \cref{tab:model-throughput} for absolute values).
	(\cref{fig:baseline_timeline}, \cref{fig:hybrid_timeline}) shows request timelines for $\text{\dedicated}_{\text{1:3}}$ and \system under a 75\% generation / 25\% embedding workload.
	Each timeline reports the number of active requests per GPU.
	\system fully utilizes all GPUs, while $\text{\dedicated}_{\text{1:3}}$ leaves some underutilized.}
	\label{fig:combined}
\end{figure*}
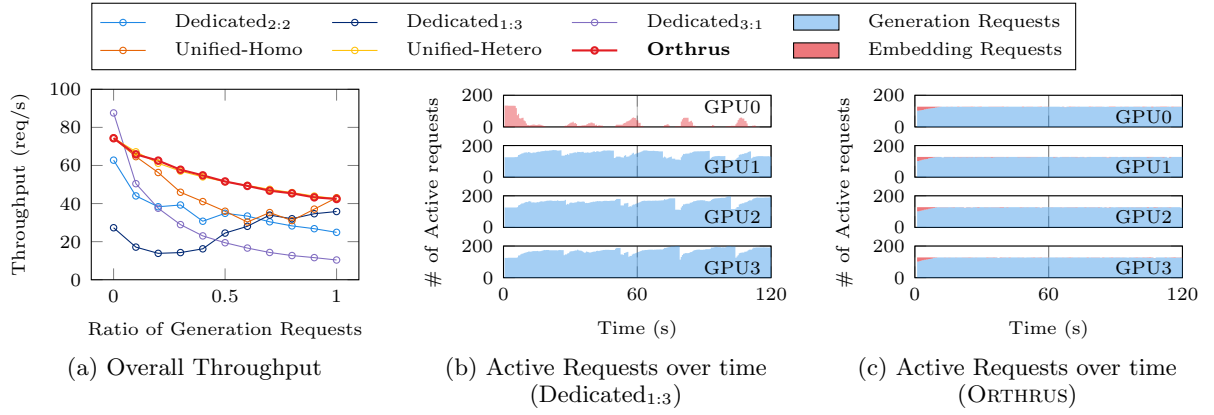

\subsection{Experimental Setup}
\label{sec:eval:setup}

\paragraph{Methods}
We consider four baselines \textbf{(Bs)}, varying model placement, batching, and scheduling (\cref{tab:baselines}).
\textbf{B1-\dedicated} runs complete embedding and generation requests on disjoint, task-specific GPU pools (e.g., $\text{\dedicated}_{\text{1:3}}$ assigns 1 GPU to embedding and 3 to generation);
\textbf{B2-\disaggpd} disaggregates generation prefill and decode across phase-specific GPU pools~\citep{zhong_distserve_2024}; embedding requests complete after pooling on the prefill GPU;
\textbf{B3-\sharedhomo} co-locates both types on the same GPU with homogeneous batching;
\textbf{B4-\sharedhetero} adds heterogeneous batching but uses \fcfs scheduling.
Finally, \textbf{\system (ours)} combines same-GPU placement, heterogeneous batching, and \ibs.
\blue{Unless otherwise specified, we report the median value across three independent runs.}

\begin{table}[t]
	\centering
	\scriptsize
	\caption{System configurations evaluated. \system is the only configuration that combines same-GPU placement, heterogeneous batching, and proportional scheduling (\ibs).}

	\vspace{-1mm}

	\label{tab:baselines}
	\resizebox{\linewidth}{!}{%
		\begin{tabular}{l|ccc}
			\toprule
			\textbf{Config.} & \makecell{\textbf{Model}                                  \\\textbf{Placement}} & \textbf{Batching} & \textbf{Scheduling} \\
			\midrule
			\dedicated       & Task-specific GPUs       & Homo            & FCFS         \\
			\disaggpd        & Phase-specific GPUs      & Homo            & FCFS         \\
			\sharedhomo      & Same GPU                 & Homo            & FCFS         \\
			\sharedhetero    & Same GPU                 & Hetero          & FCFS         \\
			\textbf{\system} & \textbf{Same GPU}        & \textbf{Hetero} & \textbf{IBS} \\
			\bottomrule
		\end{tabular}
	}
\end{table}

%
%
%
%

\paragraph{Workloads, Models, and Hardware}
We used Mistral-7B~\citep{jiang_mistral_2023} as the base model with e5-mistral-7b-instruct LoRA~\citep{wang_improving_2024} for embeddings (more models in \cref{sec:eval:generalizability}).
Synthetic workloads use 128-token embeddings and 128/512-token generation (prompt/output).
For realistic evaluation, we used Iter-RetGen~\citep{shao_enhancing_2023} on 2WikiMultihopQA~\citep{xanh2020_2wikimultihop} (avg.\ 500-token prompts, avg.\ 3000-token decodes, 1--10 documents per query).
Experiments ran on up to four A100 40GB GPUs with round-robin request distribution.

For \system, \sharedhomo, and \sharedhetero, embedding and generation requests are co-located on a single GPU using LoRA adapters.
For \dedicated, embeddings are by e5-mistral-7b-instruct; Mistral-7B serves generation.
\blue{We also study scheduling-parameter sensitivity (\cref{sec:app:parameter-sensitivity}) and chunked embedding and incremental pooling (\cref{sec:app:incremental-ablation}).} 

\subsection{High Throughput}

\label{sec:eval:throughput}

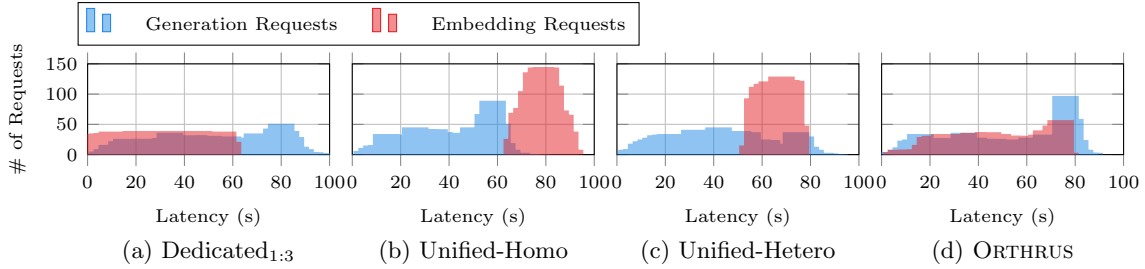
\begin{figure*}[ht]
	\centering
	\begin{tikzpicture}
    \begin{groupplot}[
        group style={
            group size=4 by 1,
            horizontal sep=3mm,
            ylabels at=edge left,
            yticklabels at=edge left,
        },
        width=0.2\textwidth,
        height=1.2cm,
        scale only axis,
        ybar,
        xmin=0, xmax=100,
        ymin=0, ymax=150,
        ytick={0, 50, 100, 150},
        grid=both,
        xlabel={Latency (s)},
        label style={font=\scriptsize},
        tick label style={font=\scriptsize},
    ]

        \nextgroupplot[
            ylabel={\# of Requests},
        ]
        \addplot+[fill=blue, draw=none, fill opacity=0.5, bar shift=1]
            table[x=bin_center, y=gen_count, col sep=comma] {figures/data/baseline_latency_histogram.csv};
        \addplot+[fill=red, draw=none, fill opacity=0.5, bar shift=1]
            table[x=bin_center, y=embed_count, col sep=comma] {figures/data/baseline_latency_histogram.csv};

        \nextgroupplot
        \addplot+[fill=blue, draw=none, fill opacity=0.5, bar shift=1]
            table[x=bin_center, y=gen_count, col sep=comma] {figures/data/turn_latency_histogram.csv};
        \addplot+[fill=red, draw=none, fill opacity=0.5, bar shift=1]
            table[x=bin_center, y=embed_count, col sep=comma] {figures/data/turn_latency_histogram.csv};

        \nextgroupplot
        \addplot+[fill=blue, draw=none, fill opacity=0.5, bar shift=1]
            table[x=bin_center, y=gen_count, col sep=comma] {figures/data/fcfs_latency_histogram.csv};
        \addplot+[fill=red, draw=none, fill opacity=0.5, bar shift=1]
            table[x=bin_center, y=embed_count, col sep=comma] {figures/data/fcfs_latency_histogram.csv};

        \nextgroupplot[
            legend style={
                font=\scriptsize,
                legend columns=-1,
                column sep=1em,
                at={(-1.0,1.2)},
                anchor=south east,
            },
        ]
        \addplot+[fill=blue, draw=none, fill opacity=0.5, bar shift=1]
            table[x=bin_center, y=gen_count, col sep=comma] {figures/data/split_latency_histogram.csv};
        \addplot+[fill=red, draw=none, fill opacity=0.5, bar shift=1]
            table[x=bin_center, y=embed_count, col sep=comma] {figures/data/split_latency_histogram.csv};
        \legend{Generation Requests, Embedding Requests}

    \end{groupplot}

    \setcounter{subfigure}{0}
    \node[below=10mm, font=\footnotesize, anchor=north] at (group c1r1.south) {%
        \refstepcounter{subfigure}\label{fig:latency_distributions:baseline}(\thesubfigure) $\text{\dedicated}_{\text{1:3}}$};
    \node[below=10mm, font=\footnotesize, anchor=north] at (group c2r1.south) {%
        \refstepcounter{subfigure}\label{fig:latency_distributions:turn}(\thesubfigure) \sharedhomo};
    \node[below=10mm, font=\footnotesize, anchor=north] at (group c3r1.south) {%
        \refstepcounter{subfigure}\label{fig:latency_distributions:fcfs}(\thesubfigure) \sharedhetero};
    \node[below=10mm, font=\footnotesize, anchor=north] at (group c4r1.south) {%
        \refstepcounter{subfigure}\label{fig:latency_distributions:split}(\thesubfigure) \system};
\end{tikzpicture}
	\vspace{-4mm}
		\caption{Latency distributions with various serving methods.
			Curves farther to the left indicate lower latencies.
			The workload consists of 1{,}000 generations
			followed by 1{,}000 embeddings, all submitted at time 0. \system achieves 9\% lower p99 generation latency than \dedicated, and up to 16\% lower embedding latency than other configurations.}
	\label{fig:latency_distributions}
\end{figure*}

We evaluate whether heterogeneous batching improves throughput compared to static GPU partitioning across varying workload ratios.

\paragraph{Controlled Workloads}
\blue{We use fixed embedding-to-generation ratios as controlled stress
	tests that isolate system behavior under known workload compositions.}
We evaluated throughput across different embedding-to-generation ratios with 512 clients over 180 seconds.
\cref{fig:throughput_ratio} reports combined throughput; \cref{tab:model-throughput} shows per-type throughput.
\dedicated deployments underutilize GPUs when workload ratios diverge from the static split, while \sharedhomo suffers under balanced workloads.
Across all evaluated workload ratios, \system achieved higher combined throughput than every static GPU partition.
\cref{fig:baseline_timeline} and \cref{fig:hybrid_timeline} illustrate that \system fully utilizes all GPUs, while $\text{\dedicated}_{\text{1:3}}$ leaves the embedding GPU underutilized.



\begin{table}[t]
	\centering
	\caption{
		E2E request throughput (req/s) on Iter-RetGen~\citep{shao_enhancing_2023} with 2WikiMultihopQA~\citep{xanh2020_2wikimultihop}.
		The workload naturally results in variable-length prompts (avg 500, max 2000 tokens), long decode lengths (avg 3000, max 4000 tokens), and multi-document retrieval (1--10 documents per query).
		\system achieves the highest throughput across all configurations.
	}
	\vspace{-1mm}
	\label{tab:rag-throughput}
	\resizebox{\linewidth}{!}{%
		\begin{tabular}{l|cc|cc|cc}
			\toprule
			method
			                        & \multicolumn{2}{c|}{1 GPU}
			                        & \multicolumn{2}{c|}{2 GPUs}
			                        & \multicolumn{2}{c}{4 GPUs}                                                   \\
			\cmidrule(lr){2-3}\cmidrule(lr){4-5}\cmidrule(lr){6-7}
			                        & Embed                       & Gen            & Embed & Gen   & Embed & Gen   \\
			\midrule
			\dedicated              & N/A                         & N/A            & 0.08  & 0.11  & 0.119 & 0.141 \\
			\sharedhomo             & 0.067                       & 0.08           & 0.126 & 0.148 & 0.252 & 0.296 \\
			\disaggpd               & N/A                         & N/A            & 0.118 & 0.14  & 0.276 & 0.327 \\
			\textbf{\system (ours)} & \textbf{0.075}              & \textbf{0.088}
			                        & \textbf{0.142}              & \textbf{0.166}
			                        & \textbf{0.288}              & \textbf{0.336}                                 \\
			\bottomrule
		\end{tabular}
	}
	\vspace{-5mm}
\end{table}

\paragraph{RAG Benchmarks}
\blue{To complement the controlled workloads, we evaluate \system on a benchmark-derived Iter-RetGen pipeline.
	We exclude vector-search and network latency to isolate model-side serving.
	Prior RAG studies report retrieval on a millisecond scale compared with
	multi-second generation~\citep{jin_ragcache_2024,wang_query2doc_2023,trivedi_interleaving_2023}.
	Each query alternates embedding, retrieval, and generation across multiple iterations,
		producing variable-length prompts and long decodes while retrieving 1--10
		documents per query. \Cref{tab:rag-throughput} shows that \system achieves the
	highest throughput at every GPU count, including the single-GPU
	setting, where disaggregation is infeasible. At four GPUs, its
	aggregate throughput is 0.624 requests/s, compared with 0.548 for
	\sharedhomo, 0.603 for \disaggpd, and 0.260 for \dedicated.}

\subsection{Low Latency with \ibs}
\label{sec:eval:starvation}

We evaluate whether \ibs reduces latency by preventing head-of-line blocking, where long-running generation requests stall shorter embedding requests under \fcfs scheduling.

\paragraph{Controlled Workloads}
We submitted 1{,}000 generation requests followed by 1{,}000 embedding requests at time 0, simulating worst-case head-of-line blocking.
Measured from submission through completion, including queueing, \system reduced p99 generation latency by 9\% over \dedicated (89\,s $\rightarrow$ 81\,s) and p99 embedding latency by 16\% over the other schedulers (87\,s $\rightarrow$ 73\,s).
In contrast, \sharedhomo and \sharedhetero both suffered from embedding requests queuing behind generations.
\blue{We further evaluate latency robustness across workload ratios and decode
	lengths in \cref{sec:app:latency-robustness}.}

\begin{table}[t]
	\centering
	\caption{
		E2E p99 latency (s) on Iter-RetGen~\citep{shao_enhancing_2023} with 2WikiMultihopQA~\citep{xanh2020_2wikimultihop}.
		We measure p99 latency per-query end-to-end (including all retrieval iterations).
		\system achieves the lowest latency across all GPU counts.
	}
	\vspace{-1mm}
	\label{tab:rag-latency}
	\small
	\begin{tabular}{l|ccc}
		\toprule
		\multirow{2}{*}{Method} & \multicolumn{3}{c}{E2E p99 Latency (s)}                                   \\
		\cmidrule(lr){2-4}
		                        & 1 GPU                                   & 2 GPUs         & 4 GPUs         \\
		\midrule
		$\text{\dedicated}$     & N/A                                     & 960.1          & 1342.0         \\
		\disaggpd               & N/A                                     & 602.6          & 617.0          \\
		\sharedhomo             & 673.4                                   & 676.0          & 700.6          \\
		\textbf{\system (ours)} & \textbf{573.9}                          & \textbf{600.5} & \textbf{593.4} \\
		\bottomrule
	\end{tabular}
\end{table}

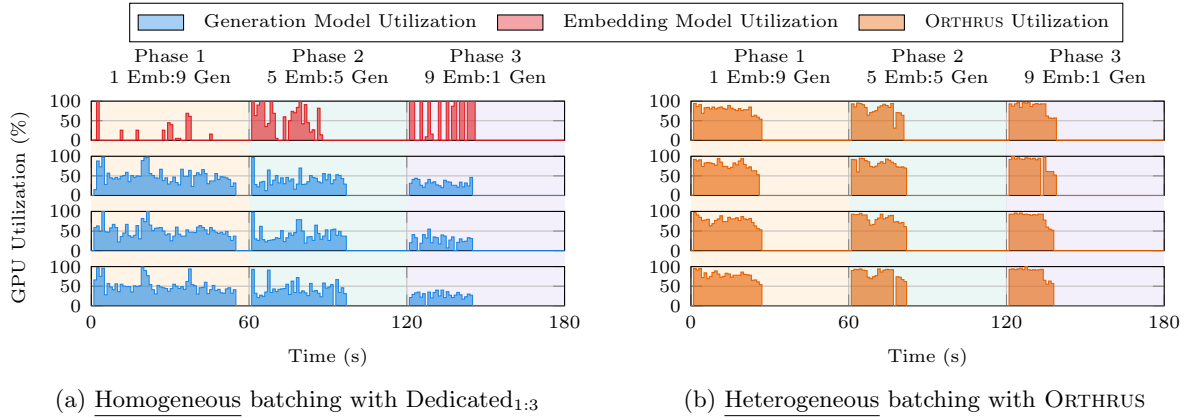
\begin{figure*}[th]
	\centering
	\begin{subfigure}[b]{0.49\linewidth}
		\centering
		\begin{tikzpicture}
  \begin{groupplot}[
      group style={
        group size=1 by 4,
        vertical sep=6pt,
        x descriptions at=edge bottom,
      },
      width=\linewidth,
      height=2.1cm,
      xmin=0, xmax=180,
      ymin=0, ymax=100,
      xlabel={Time (s)},
      grid=both,
      tick label style={font=\scriptsize},
      label style={font=\scriptsize},
      set layers,
      legend style={
        font=\scriptsize,
        legend columns=3,
        at={(0.08,2.5)},
        anchor=south west,
        column sep=0.5em,
      },
    ]
    \pgfplotsset{every axis title/.append style={at={(1,0.9)},
    font=\scriptsize, anchor=north east}}

    \nextgroupplot[
      xtick={0, 60,120, 180},
      ytick={0, 50, 100},
    ]

    \addlegendimage{area legend, draw=none, fill=blue, fill opacity=0.40}
    \addlegendentry{Generation Model Utilization}
    \addlegendimage{area legend, draw=none, fill=red, fill opacity=0.40}
    \addlegendentry{Embedding Model Utilization}
    \addlegendimage{area legend, draw=none, fill=orange, fill opacity=0.40}
    \addlegendentry{\system Utilization}

    \coordinate (baselineShadeTopXZero) at (axis
    cs:0,\pgfkeysvalueof{/pgfplots/ymax});
    \coordinate (baselineShadeTopXSixty) at (axis
    cs:60,\pgfkeysvalueof{/pgfplots/ymax});
    \coordinate (baselineShadeTopXOneTwenty) at (axis
    cs:120,\pgfkeysvalueof{/pgfplots/ymax});
    \coordinate (baselineShadeTopXOneSixty) at (axis
    cs:180,\pgfkeysvalueof{/pgfplots/ymax});
    \path (baselineShadeTopXZero) -- (baselineShadeTopXSixty)
    coordinate[pos=0.5] (baselineShadeLabelPhaseOne);
    \path (baselineShadeTopXSixty) -- (baselineShadeTopXOneTwenty)
    coordinate[pos=0.5] (baselineShadeLabelPhaseTwo);
    \path (baselineShadeTopXOneTwenty) -- (baselineShadeTopXOneSixty)
    coordinate[pos=0.5] (baselineShadeLabelPhaseThree);
    \addplot[name path=baselineZeroZero, draw=none] {0};
    \addplot+[
      name path=baselinePathZero,
      const plot, no marks, draw=none,
      restrict expr to domain={\thisrow{gpu_index}}{0:0}
    ] table[
      col sep=comma, x=t, y=util_percent
    ] {figures/data/baseline_util_timeline_all.csv};
    \addplot[
      fill=red,
      fill opacity=0.6,
      draw=none
    ] fill between[of=baselinePathZero and baselineZeroZero];
    \addplot+[
      const plot, no marks, draw=red,
      restrict expr to domain={\thisrow{gpu_index}}{0:0}
    ] table[
      col sep=comma, x=t, y=util_percent
    ] {figures/data/baseline_util_timeline_all.csv};

    \nextgroupplot[
      xtick={0, 60,120, 180},
      ytick={0, 50, 100},
    ]
    \addplot[name path=baselineZeroOne, draw=none] {0};
    \addplot+[
      name path=baselinePathOne,
      const plot, no marks, draw=none,
      restrict expr to domain={\thisrow{gpu_index}}{1:1}
    ] table[
      col sep=comma, x=t, y=util_percent
    ] {figures/data/baseline_util_timeline_all.csv};
    \addplot[
      fill=blue,
      fill opacity=0.6,
      draw=none
    ] fill between[of=baselinePathOne and baselineZeroOne];
    \addplot+[
      const plot, no marks, draw=blue,
      restrict expr to domain={\thisrow{gpu_index}}{1:1}
    ] table[
      col sep=comma, x=t, y=util_percent
    ] {figures/data/baseline_util_timeline_all.csv};

    \nextgroupplot[
      ylabel={GPU Utilization (\%)},
      every axis y label/.append style={at=(ticklabel cs:1.1)},
      xtick={0, 60,120, 180},
      ytick={0, 50, 100},
    ]
    \addplot[name path=baselineZeroTwo, draw=none] {0};
    \addplot+[
      name path=baselinePathTwo,
      const plot, no marks, draw=none,
      restrict expr to domain={\thisrow{gpu_index}}{2:2}
    ] table[
      col sep=comma, x=t, y=util_percent
    ] {figures/data/baseline_util_timeline_all.csv};
    \addplot[
      fill=blue,
      fill opacity=0.6,
      draw=none
    ] fill between[of=baselinePathTwo and baselineZeroTwo];
    \addplot+[
      const plot, no marks, draw=blue,
      restrict expr to domain={\thisrow{gpu_index}}{2:2}
    ] table[
      col sep=comma, x=t, y=util_percent
    ] {figures/data/baseline_util_timeline_all.csv};

    \nextgroupplot[
      xlabel={Time (s)},
      xtick={0,60,120,180},
      ytick={0, 50, 100},
    ]
    \addplot[name path=baselineZeroThree, draw=none] {0};
    \addplot+[
      name path=baselinePathThree,
      const plot, no marks, draw=none,
      restrict expr to domain={\thisrow{gpu_index}}{3:3}
    ] table[
      col sep=comma, x=t, y=util_percent
    ] {figures/data/baseline_util_timeline_all.csv};
    \addplot[
      fill=blue,
      fill opacity=0.6,
      draw=none
    ] fill between[of=baselinePathThree and baselineZeroThree];
    \addplot+[
      const plot, no marks, draw=blue,
      restrict expr to domain={\thisrow{gpu_index}}{3:3}
    ] table[
      col sep=comma, x=t, y=util_percent
    ] {figures/data/baseline_util_timeline_all.csv};
    \coordinate (baselineShadeBottomXZero) at (axis
    cs:0,\pgfkeysvalueof{/pgfplots/ymin});
    \coordinate (baselineShadeBottomXSixty) at (axis
    cs:60,\pgfkeysvalueof{/pgfplots/ymin});
    \coordinate (baselineShadeBottomXOneTwenty) at (axis
    cs:120,\pgfkeysvalueof{/pgfplots/ymin});
    \coordinate (baselineShadeBottomXOneSixty) at (axis
    cs:180,\pgfkeysvalueof{/pgfplots/ymin});

  \end{groupplot}
  \begin{pgfonlayer}{axis background}
    \fill[phaseone, fill opacity=0.5, draw=none]
    (baselineShadeTopXZero) --
    (baselineShadeTopXSixty) --
    (baselineShadeBottomXSixty) --
    (baselineShadeBottomXZero) --
    cycle;
    \fill[phasetwo, fill opacity=0.5, draw=none]
    (baselineShadeTopXSixty) --
    (baselineShadeTopXOneTwenty) --
    (baselineShadeBottomXOneTwenty) --
    (baselineShadeBottomXSixty) --
    cycle;
    \fill[phasethree, fill opacity=0.5, draw=none]
    (baselineShadeTopXOneTwenty) --
    (baselineShadeTopXOneSixty) --
    (baselineShadeBottomXOneSixty) --
    (baselineShadeBottomXOneTwenty) --
    cycle;
  \end{pgfonlayer}
  \node[anchor=south, font=\scriptsize] at
  ([yshift=3pt]baselineShadeLabelPhaseOne) {\shortstack{Phase~1 \\ 1
  Emb:9 Gen}};
  \node[anchor=south, font=\scriptsize] at
  ([yshift=3pt]baselineShadeLabelPhaseTwo) {\shortstack{Phase~2 \\ 5
  Emb:5 Gen}};
  \node[anchor=south, font=\scriptsize] at
  ([yshift=3pt]baselineShadeLabelPhaseThree) {\shortstack{Phase~3
  \\ 9 Emb:1 Gen}};
\end{tikzpicture}
		\vspace{-10mm}
		\caption{\underline{Homogeneous} batching with
		$\text{\dedicated}_{\text{1:3}}$}
		\label{fig:baseline_utilization}
	\end{subfigure}
	\hfill
	\begin{subfigure}[b]{0.49\linewidth}
		\centering
		\begin{tikzpicture}
  \begin{groupplot}[
      group style={
        group size=1 by 4,
        vertical sep=6pt,
        x descriptions at=edge bottom,
      },
      width=\linewidth,
      height=2.1cm,
      xmin=0, xmax=180,
      ymin=0, ymax=100,
      xlabel={Time (s)},
      grid=both,
      tick label style={font=\scriptsize},
      label style={font=\scriptsize},
      set layers,
    ]

    \nextgroupplot[
      xtick={0, 60,120,180},
      ytick={0, 50, 100},
    ]
    \coordinate (shadeTopXZero) at (axis cs:0,\pgfkeysvalueof{/pgfplots/ymax});
    \coordinate (shadeTopXSixty) at (axis
    cs:60,\pgfkeysvalueof{/pgfplots/ymax});
    \coordinate (shadeTopXOneTwenty) at (axis
    cs:120,\pgfkeysvalueof{/pgfplots/ymax});
    \coordinate (shadeTopXOneSixty) at (axis
    cs:180,\pgfkeysvalueof{/pgfplots/ymax});
    \path (shadeTopXZero) -- (shadeTopXSixty) coordinate[pos=0.5]
    (shadeLabelPhaseOne);
    \path (shadeTopXSixty) -- (shadeTopXOneTwenty)
    coordinate[pos=0.5] (shadeLabelPhaseTwo);
    \path (shadeTopXOneTwenty) -- (shadeTopXOneSixty)
    coordinate[pos=0.5] (shadeLabelPhaseThree);
    \addplot[name path=splitZeroZero, draw=none] {0};
    \addplot+[
      name path=splitPathZero,
      const plot, no marks, draw=none,
      restrict expr to domain={\thisrow{gpu_index}}{0:0}
    ] table[
      col sep=comma, x=t, y=util_percent
    ] {figures/data/hybrid_split_util_timeline_all.csv};
    \addplot[
      fill=orange,
      fill opacity=0.6,
      draw=none
    ] fill between[of=splitPathZero and splitZeroZero];
    \addplot+[
      const plot, no marks, draw=orange,
      restrict expr to domain={\thisrow{gpu_index}}{0:0}
    ] table[
      col sep=comma, x=t, y=util_percent
    ] {figures/data/hybrid_split_util_timeline_all.csv};

    \nextgroupplot[xtick={0, 60,120,180},
      ytick={0, 50, 100},
    ]
    \addplot[name path=splitZeroOne, draw=none] {0};
    \addplot+[
      name path=splitPathOne,
      const plot, no marks, draw=none,
      restrict expr to domain={\thisrow{gpu_index}}{1:1}
    ] table[
      col sep=comma, x=t, y=util_percent
    ] {figures/data/hybrid_split_util_timeline_all.csv};
    \addplot[
      fill=orange,
      fill opacity=0.6,
      draw=none
    ] fill between[of=splitPathOne and splitZeroOne];
    \addplot+[
      const plot, no marks, draw=orange,
      restrict expr to domain={\thisrow{gpu_index}}{1:1}
    ] table[
      col sep=comma, x=t, y=util_percent
    ] {figures/data/hybrid_split_util_timeline_all.csv};

    \nextgroupplot[
      xtick={0, 60,120,180},
      ytick={0, 50, 100},
    ]
    \addplot[name path=splitZeroTwo, draw=none] {0};
    \addplot+[
      name path=splitPathTwo,
      const plot, no marks, draw=none,
      restrict expr to domain={\thisrow{gpu_index}}{2:2}
    ] table[
      col sep=comma, x=t, y=util_percent
    ] {figures/data/hybrid_split_util_timeline_all.csv};
    \addplot[
      fill=orange,
      fill opacity=0.6,
      draw=none
    ] fill between[of=splitPathTwo and splitZeroTwo];
    \addplot+[
      const plot, no marks, draw=orange,
      restrict expr to domain={\thisrow{gpu_index}}{2:2}
    ] table[
      col sep=comma, x=t, y=util_percent
    ] {figures/data/hybrid_split_util_timeline_all.csv};

    \nextgroupplot[
      xlabel={Time (s)},
      xtick={0, 60, 120, 180},
      ytick={0, 50, 100},
    ]
    \addplot[name path=splitZeroThree, draw=none] {0};
    \addplot+[
      name path=splitPathThree,
      const plot, no marks, draw=none,
      restrict expr to domain={\thisrow{gpu_index}}{3:3}
    ] table[
      col sep=comma, x=t, y=util_percent
    ] {figures/data/hybrid_split_util_timeline_all.csv};
    \addplot[
      fill=orange,
      fill opacity=0.6,
      draw=none
    ] fill between[of=splitPathThree and splitZeroThree];
    \addplot+[
      const plot, no marks, draw=orange,
      restrict expr to domain={\thisrow{gpu_index}}{3:3}
    ] table[
      col sep=comma, x=t, y=util_percent
    ] {figures/data/hybrid_split_util_timeline_all.csv};
    \coordinate (shadeBottomXZero) at (axis
    cs:0,\pgfkeysvalueof{/pgfplots/ymin});
    \coordinate (shadeBottomXSixty) at (axis
    cs:60,\pgfkeysvalueof{/pgfplots/ymin});
    \coordinate (shadeBottomXOneTwenty) at (axis
    cs:120,\pgfkeysvalueof{/pgfplots/ymin});
    \coordinate (shadeBottomXOneSixty) at (axis
    cs:180,\pgfkeysvalueof{/pgfplots/ymin});

  \end{groupplot}
  \begin{pgfonlayer}{axis background}
    \fill[phaseone, fill opacity=0.5, draw=none]
    (shadeTopXZero) --
    (shadeTopXSixty) --
    (shadeBottomXSixty) --
    (shadeBottomXZero) --
    cycle;
    \fill[phasetwo, fill opacity=0.5, draw=none]
    (shadeTopXSixty) --
    (shadeTopXOneTwenty) --
    (shadeBottomXOneTwenty) --
    (shadeBottomXSixty) --
    cycle;
    \fill[phasethree, fill opacity=0.5, draw=none]
    (shadeTopXOneTwenty) --
    (shadeTopXOneSixty) --
    (shadeBottomXOneSixty) --
    (shadeBottomXOneTwenty) --
    cycle;
  \end{pgfonlayer}
  \node[anchor=south, font=\scriptsize] at
  ([yshift=3pt]baselineShadeLabelPhaseOne) {\shortstack{Phase~1 \\ 1
  Emb:9 Gen}};
  \node[anchor=south, font=\scriptsize] at
  ([yshift=3pt]baselineShadeLabelPhaseTwo) {\shortstack{Phase~2 \\ 5
  Emb:5 Gen}};
  \node[anchor=south, font=\scriptsize] at
  ([yshift=3pt]baselineShadeLabelPhaseThree) {\shortstack{Phase~3
  \\ 9 Emb:1 Gen}};
\end{tikzpicture}
		\vspace{-10mm}
		\caption{\underline{Heterogeneous} batching with \system}
		\label{fig:split_utilization}
	\end{subfigure}
	\caption{
	GPU utilization under three workload phases.
	$\text{\dedicated}_{\text{1:3}}$ (\cref{fig:baseline_utilization}) underutilizes GPUs (avg.\ 38\%), while
	\system (\cref{fig:split_utilization}) sustains 79\% utilization and completes each phase 43\% faster.
	}
	\label{fig:utilization_comparison}
\end{figure*}

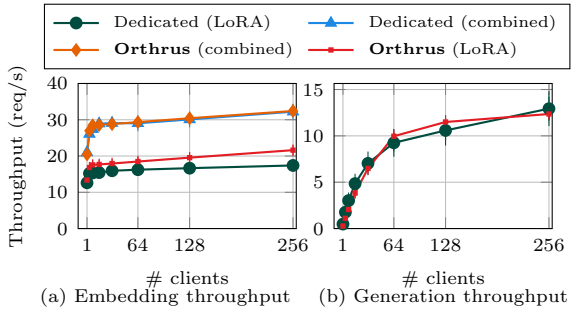
\begin{figure}[ht]
	\centering
\begin{tikzpicture}
\begin{groupplot}[
    group style={
        group size=2 by 1,
        horizontal sep=5mm,
        ylabels at=edge left,
    },
    width=0.58\linewidth,
    height=35mm,
    xlabel={\# clients},
    xmin=-10, xmax=260,
    ymin=0,
    xtick={1, 64, 128, 256},
    xmajorgrids, ymajorgrids,
    label style={
      font=\scriptsize,
    },
    tick label style={font=\scriptsize},
    every mark/.append style={scale=0.5},
    title style={at={(0.5,-0.45)}, anchor=north, font=\scriptsize},
]

\nextgroupplot[
    title={(a) Embedding throughput},
    title style={at={(0.4,-0.45)}, anchor=north, font=\scriptsize},
    ylabel={Throughput (req/s)},
    ymax=40,
    legend style={
        legend cell align={left},
        font=\tiny,
        legend columns=2,
        column sep=0.3em,
        at={(1.05, 1.55)},
        anchor=north,
    },
]

\addplot+[mark=*, thick, color=green,
          error bars/.cd, y dir=both, y explicit, error mark options={draw=green, mark size=2pt}]
coordinates {
  (1,12.616667) +- (0,0.0631)
  (4,15.200000) +- (0,0.190)
  (8,15.333333) +- (0,0.0727)
  (16,15.433333) +- (0,0.0747)
  (32,15.933333) +- (0,0.0747)
  (64,16.23333) +- (0,0.0714)
  (128,16.650000) +- (0,0.168)
  (256,17.39999) +- (0,0.174)
};
\addlegendentry{\dedicated~(LoRA)}

\addplot+[mark=triangle*, thick, color=blue,
          error bars/.cd, y dir=both, y explicit, error mark options={draw=blue, mark size=2pt}]
coordinates {
  (1,20.983333) +- (0,0.141421)
  (4,26.066667) +- (0,0.094281)
  (8,27.400000) +- (0,0.094281)
  (16,28.933333) +- (0,0.188562)
  (32,29.066667) +- (0,0.377123)
  (64,29.025000) +- (0,0.718891)
  (128,30.091667) +- (0,0.718892)
  (256,32.225000) +- (0,0.718891)
};
\addlegendentry{\dedicated~(combined)}

\addplot+[mark=diamond*, thick, color=orange,
          error bars/.cd, y dir=both, y explicit, error mark options={draw=orange, mark size=2pt}]
coordinates {
  (1,20.433333) +- (0,0.07022)
  (4,26.950000) +- (0,0.07348)
  (8,28.466667) +- (0,0.07423)
  (16,28.550000) +- (0,0.07428)
  (32,28.816667) +- (0,0.08441)
  (64,29.350000) +- (0,0.09468)
  (128,30.416667) +- (0,0.1521)
  (256,32.416667) +- (0,0.1621)
};
\addlegendentry{\textbf{\system}~(combined)}

\addplot+[mark=square*, thick, color=red, mark options={fill=red, scale=0.3},
          error bars/.cd, y dir=both, y explicit, error mark options={draw=red, mark size=2pt}]
coordinates {
  (1,13.383333) +- (0,0.086603)
  (4,16.827778) +- (0,0.067358)
  (8,17.550000) +- (0,0.115470)
  (16,17.683333) +- (0,0.115470)
  (32,17.927778) +- (0,0.153960)
  (64,18.483333) +- (0,0.115470)
  (128,19.550000) +- (0,0.115410)
  (256,21.616666) +- (0,0.115470)
};
\addlegendentry{\textbf{\system}~(LoRA)}

\nextgroupplot[
    title={(b) Generation throughput},
]

\addplot+[mark=*, thick, color=green,
          error bars/.cd, y dir=both, y explicit, error mark options={draw=green, mark size=2pt}]
coordinates {
  (1,0.483333) +- (0,0.050)
  (4,1.766667) +- (0,0.177)
  (8,3.016667) +- (0,0.302)
  (16,4.850000) +- (0,0.485)
  (32,7.033333) +- (0,0.703)
  (64,9.250000) +- (0,0.925)
  (128,10.583333) +- (0,1.058)
  (256,12.933333) +- (0,1.293)
};

\addplot+[mark=square*, thick, color=red, mark options={fill=red, scale=0.3},
          error bars/.cd, y dir=both, y explicit, error mark options={draw=red, mark size=2pt}]
coordinates {
  (1,0.277778) +- (0,0.009622)
  (4,1.105556) +- (0,0.019245)
  (8,2.050000) +- (0,0.028868)
  (16,3.827778) +- (0,0.038490)
  (32,6.488889) +- (0,0.105847)
  (64,9.961111) +- (0,0.153960)
  (128,11.494445) +- (0,0.125093)
  (256,12.355556) +- (0,0.009623)
};

\end{groupplot}

\end{tikzpicture}

	\caption{
		Overhead analysis of \system on a single GPU.
		At $\geq 64$ clients, \system reaches similar saturated generation
		throughput to dedicated vLLM. Embedding throughput is comparable to or
		higher than the corresponding dedicated configuration; the gap between the
		combined-model and LoRA curves reflects adapter overhead.}
	\label{fig:overhead}
\end{figure}

\paragraph{RAG Benchmarks}
\cref{tab:rag-latency} shows that \system achieved the lowest latency across all GPU counts:
14.8\% lower than \sharedhomo on 1 GPU (573.9s vs.\ 673.4s),
comparable to \disaggpd on 2 GPUs (600.5s vs.\ 602.6s),
and 55.8\% lower than \dedicated on 4 GPUs (593.4s vs.\ 1342.0s).
\system avoids both static partitioning bottlenecks and cross-GPU KV cache transfer.

\subsection{High GPU Utilization}
\label{sec:eval:load_balance}

We evaluate whether heterogeneous batching improves GPU utilization over static partitioning under shifting workload ratios.
Across three phases (10\%/50\%/90\% embedding), \system achieved 79\% average utilization versus 38\% for $\text{\dedicated}_{\text{1:3}}$, completing each phase 43\% faster by balancing load across GPUs.
Static partitioning underperformed: the dedicated embedding GPU was underutilized during generation-heavy phases and overloaded during embedding-heavy phases.

\subsection{Minimal Overhead}
\label{sec:eval:tax}

\Cref{fig:overhead} compares embedding- and generation-only workloads on a single GPU with 1--256 clients.
\system reaches dedicated vLLM generation throughput once the workload saturates ($\geq64$ clients), yet it trails at lower concurrency.
Its embedding throughput is comparable to or higher than the corresponding dedicated configuration; the lower absolute throughput of both LoRA curves reflects adapter overhead.
These results show that heterogeneous batching preserves saturated per-task throughput while enabling GPU sharing.
We study the contribution of chunked embedding and incremental pooling in
\cref{sec:app:incremental-ablation}.

\begin{table}[t]
	\centering
	\caption{
		Generalizability analysis comparing absolute throughput (requests/sec) across different models and request mixes.
		We report embedding and generation throughput separately for three ratios (embedding:generation).
		\dedicated runs separate models on a 1:3 GPU split. \system co-locates both tasks on every GPU.
	}
	\vspace{-1mm}
	\resizebox{\linewidth}{!}{%
		\label{tab:model-throughput}
		\begin{tabular}{llc|cc|cc}
			\toprule
			Model & LoRA                                   & Ratio & Embed                           & Gen                                  & Embed          & Gen            \\
			      &                                        &       & \multicolumn{2}{c|}{\dedicated} & \multicolumn{2}{c}{\textbf{\system}}                                   \\
			\midrule
			\multirow{3}{*}{Mistral 7B}
			      & \multirow{3}{*}{e5-mistral}
			      & 9:1                                    & 15.64 & 1.49                            & \textbf{58.80}                       & \textbf{7.00}                   \\
			      &                                        & 5:5   & 12.54                           & 11.97                                & \textbf{25.32} & \textbf{26.24} \\
			      &                                        & 1:9   & 5.10                            & 29.56                                & 4.20           & \textbf{39.10} \\
			\midrule
			\multirow{3}{*}{Qwen2 7B}
			      & \multirow{3}{*}{\makecell[l]{Synthetic                                                                                                                    \\ Rank 32}}
			      & 9:1                                    & 13.23 & 1.72                            & \textbf{40.90}                       & \textbf{4.40}                   \\
			      &                                        & 5:5   & 10.72                           & 10.08                                & \textbf{26.80} & \textbf{25.93} \\
			      &                                        & 1:9   & 2.81                            & 27.76                                & \textbf{6.20}  & \textbf{48.23} \\
			\midrule
			\multirow{3}{*}{LLaMA3.1 8B}
			      & \multirow{3}{*}{\makecell[l]{Synthetic                                                                                                                    \\ Rank 64}}
			      & 9:1                                    & 13.43 & 1.32                            & \textbf{56.46}                       & \textbf{5.90}                   \\
			      &                                        & 5:5   & 10.18                           & 9.95                                 & \textbf{34.60} & \textbf{37.50} \\
			      &                                        & 1:9   & 3.06                            & 25.63                                & \textbf{4.23}  & \textbf{47.32} \\
			\bottomrule
		\end{tabular}}
	\vspace{-2mm}
\end{table}

%

\subsection{Generalization}
\label{sec:eval:generalizability}


We evaluate whether heterogeneous batching generalizes across model architectures and LoRA adapter ranks.
We tested Mistral 7B, Qwen2 7B~\citep{yang_qwen2_2024}, and LLaMA3.1 8B~\citep{grattafiori_llama_2024} with LoRA ranks up to 64 and embedding-to-generation ratios of 9:1, 5:5, and 1:9.
As shown in \cref{tab:model-throughput}, \system achieved higher combined embedding-plus-generation throughput than $\text{\dedicated}_{\text{1:3}}$ across all configurations.
Per-type throughput was also generally higher; the exception was Mistral 7B at 1:9, where embedding throughput decreased from 5.10 to 4.20 requests/s while generation throughput increased from 29.56 to 39.10 requests/s.

\blue{\paragraph{Separate Models.}
	We also evaluate \system with separate models (OPT-1.3B for generation and GTR-T5-XL for embeddings) on a single
	A40 GPU. \system co-serves both models with a 0.9 GPU-memory cap, while the
	baseline uses two vLLM instances capped at 0.45 each. We use the
	Iter-RetGen-style 2WikiMultihopQA workload from \cref{sec:eval:throughput}.
	}

\begin{table}[t]
	\centering
	\footnotesize
	\setlength{\tabcolsep}{2.1pt}
	\caption{\blue{Separate-model performance on one A40 GPU. Gain denotes a
			throughput increase or latency reduction relative to 2$\times$vLLM;
			positive values favor \system.}}
	\label{tab:separate-model}
	\begin{tabular}{@{}llrrr@{}}
		\toprule
		Metric & Workload & 2$\times$vLLM   & \system         & Gain             \\
		\midrule
		\multirow{2}{*}{\makecell[l]{Throughput                                  \\(req/s)}}
		       & Emb.     & 29.7            & \textbf{36.5}   & \textbf{+22.9\%} \\
		       & Gen.     & 28.2            & \textbf{33.9}   & \textbf{+20.2\%} \\
		\addlinespace[1pt]
		\midrule
		\multirow{2}{*}{\makecell[l]{Avg. latency                                \\(ms)}}
		       & Emb.     & 187.0           & \textbf{151.8}  & \textbf{+18.8\%} \\
		       & Gen.     & 4337.1          & \textbf{3561.5} & \textbf{+17.9\%} \\
		\addlinespace[1pt]
		\midrule
		\multirow{2}{*}{\makecell[l]{p95 latency                                 \\(ms)}}
		       & Emb.     & 421.4           & \textbf{318.6}  & \textbf{+24.4\%} \\
		       & Gen.     & \textbf{4875.3} & 5059.7          & -3.8\%           \\
		\bottomrule
	\end{tabular}
\end{table}

\blue{\system improves embedding and generation throughput by 22.9\% and
	20.2\%, respectively, and reduces both average latencies and embedding p95
	latency. The 3.8\% increase in generation p95 latency is outweighed by the
	17.9\% reduction in mean generation latency.}

%


%

\section{Conclusion}
\label{sec:con}

This work presents
\system, a serving system that co-serves embedding and generation within a
unified model runner through heterogeneous batching.
Unlike existing systems,
\system integrates embedding into the same scheduling and execution path
as generation.
By combining chunked embedding with incremental pooling and workload-aware scheduling,
\system achieves higher throughput and lower latency under mixed workloads,
enabling efficient serving for knowledge-intensive LLM applications.

\newpage

\section*{Limitations}
\label{sec:limitations}

\blue{Our evaluation scales \system by running one model replica per GPU, with
up to four A100 GPUs. Deployments with more replicas may exhibit different
load-balancing behavior and workload skew across GPUs, which can affect
utilization and tail latency. Thus, our experiments do not establish how the
observed gains scale beyond four GPUs.}

\blue{\system does not currently implement support for a single model sharded
across multiple GPUs. The separate-model experiment likewise considers only
two models that fit together on one A40. Models that require tensor or pipeline
parallelism introduce communication and synchronization within each iteration,
which may interact differently with heterogeneous batching and scheduling.}

\blue{Our experiments use open-weight models with fewer than 10 billion
parameters. Larger models have greater memory demands and may exhibit different
compute-to-memory ratios and batch-capacity constraints. Although the results
are consistent across the tested model families, they do not establish that
\system provides the same throughput and latency benefits for larger models.}

\section*{Acknowledgments}

This work was supported in part by the IBM-Illinois Discovery Accelerator Institute and by the National Science Foundation under grants \#2312561 and \#2440498. We used Delta and DeltaAI at the National Center for Supercomputing Applications (NCSA) through allocation CIS240661 from the ACCESS program, which is supported by NSF grants \#2138259, \#2138286, \#2138307, \#2137603, and \#2138296.

\bibliography{iclr2026_conference}

\clearpage
\appendix
\section{Additional Experiments}

\subsection{Pooling Equivalence}

\blue{We test whether incremental pooling reproduces single-pass embeddings
across pooling heads and chunk sizes. On 250 samples spanning a range of input
lengths, we compare paired outputs for mean, CLS, and weighted-mean pooling at
chunk sizes 256, 512, and 1024 using cosine similarity and relative $L_2$
error. In \cref{tab:pooling-equivalence}, minimum cosine similarity exceeds
0.9999 in all nine configurations, while both mean and p99 relative $L_2$
errors remain below 1\%. These small discrepancies reflect
floating-point reduction-order effects rather than an algorithmic
approximation.}

\begin{table}[h]
	\centering
	\caption{Equivalence between chunked and single-pass embeddings.
		$C$ is the chunk size; $L_2$ errors are relative.}
	\label{tab:pooling-equivalence}
	\scriptsize
	\setlength{\tabcolsep}{1.4pt}
	\begin{tabular}{l r rrr rr}
		\toprule
		\multirow{2}{*}{Pooling} & \multirow{2}{*}{$C$}
		& \multicolumn{3}{c}{Cosine similarity}
		& \multicolumn{2}{c}{Relative $L_2$} \\
		\cmidrule(lr){3-5}\cmidrule(lr){6-7}
		& & Min & Mean & p50 & Mean & p99 \\
		\midrule
		\multirow{3}{*}{Mean}
		& 256  & .9999512 & .9999844 & .9999864 & .005527 & .008995 \\
		& 512  & .9999555 & .9999861 & .9999885 & .005176 & .009478 \\
		& 1024 & .9999484 & .9999877 & .9999881 & .004820 & .008156 \\
		\midrule
		\multirow{3}{*}{CLS}
		& 256  & .9999894 & .9999912 & .9999912 & .005072 & .005113 \\
		& 512  & .9999837 & .9999887 & .9999892 & .004908 & .005479 \\
		& 1024 & .9999920 & .9999930 & .9999920 & .003692 & .004236 \\
		\midrule
		\multirow{3}{*}{\makecell[l]{Weighted\\mean}}
		& 256  & .9999456 & .9999882 & .9999907 & .004650 & .009628 \\
		& 512  & .9999339 & .9999887 & .9999918 & .004493 & .009927 \\
		& 1024 & .9999452 & .9999905 & .9999915 & .004215 & .007636 \\
		\bottomrule
	\end{tabular}
\end{table}

\subsection{Scheduling-Parameter Sensitivity}
\label{sec:app:parameter-sensitivity}

\begin{figure}[th]
	\centering
	\begin{tikzpicture}
		\begin{groupplot}[
			group style={
				group size=2 by 2,
				horizontal sep=5mm,
				vertical sep=1mm,
				ylabels at=edge left,
				yticklabels at=edge left,
			},
			width=0.31\linewidth,
			height=14mm,
			scale only axis,
			xmode=log,
			log basis x=2,
			log ticks with fixed point,
			xmajorgrids,
			ymajorgrids,
			grid style={draw=gray!25},
			label style={font=\tiny, align=center},
			tick label style={font=\tiny},
			every axis plot/.append style={thick, mark size=1.5pt},
		]

		\nextgroupplot[
			ylabel={Throughput\\(req/s)},
			xmin=110, xmax=1200,
			xtick={128,256,512,1024},
			xticklabels=\empty,
			ymin=0, ymax=3,
			ytick={0,1,2,3},
			legend to name=parameter-sensitivity-legend,
			legend style={
				legend cell align={left},
				font=\tiny,
				legend columns=2,
				column sep=0.3em,
			},
		]
		\addplot+[color=combinedcolor, mark=diamond*] coordinates {
			(128,2.6167) (256,2.6500) (512,2.6500) (1024,2.6208)
		};
		\addlegendentry{Combined throughput}
		\addlegendimage{color=generationcolor, mark=triangle*, thick}
		\addlegendentry{Gen. p99}
		\addlegendimage{color=embeddingcolor, mark=square*, thick}
		\addlegendentry{Embed. p99}

		\nextgroupplot[
			xmin=1750, xmax=9500,
			xtick={2048,4096,8192},
			xticklabels=\empty,
			ymin=0, ymax=3,
			ytick={0,1,2,3},
		]
		\addplot+[color=combinedcolor, mark=diamond*] coordinates {
			(2048,2.6125) (4096,2.6208) (8192,2.6375)
		};

		\nextgroupplot[
			ylabel={Gen. p99\\(s)},
			xmin=110, xmax=1200,
			xtick={128,256,512,1024},
			xticklabels={128,256,512,1024},
			ymin=0, ymax=4.5,
			ytick={0,1,2,3,4},
		]
		\addplot+[color=generationcolor, mark=triangle*] coordinates {
			(128,4.16107) (256,4.07663) (512,4.07746) (1024,4.14144)
		};

		\nextgroupplot[
			xmin=1750, xmax=9500,
			xtick={2048,4096,8192},
			xticklabels={2048,4096,8192},
			ymin=0, ymax=4.5,
			ytick={0,1,2,3,4},
		]
		\addplot+[color=generationcolor, mark=triangle*] coordinates {
			(2048,4.12226) (4096,4.05796) (8192,4.06490)
		};
		\end{groupplot}

		\begin{axis}[
			at={(group c1r2.south west)},
			anchor=south west,
			width=0.31\linewidth,
			height=14mm,
			scale only axis,
			xmode=log,
			log basis x=2,
			xmin=110, xmax=1200,
			axis x line=none,
			axis y line=none,
			ymin=0, ymax=120,
			clip mode=individual,
		]
		\addplot+[color=embeddingcolor, mark=square*, thick, mark size=1.5pt] coordinates {
			(128,89.74) (256,89.20) (512,83.58) (1024,98.77)
		};
		\end{axis}

		\begin{axis}[
			at={(group c2r2.south west)},
			anchor=south west,
			width=0.31\linewidth,
			height=14mm,
			scale only axis,
			xmode=log,
			log basis x=2,
			xmin=1750, xmax=9500,
			axis x line=none,
			axis y line*=right,
			ylabel={Embed. p99\\(ms)},
			ymin=0, ymax=120,
			ytick={0,40,80,120},
			label style={font=\tiny, align=center},
			tick label style={font=\tiny},
			clip mode=individual,
		]
		\addplot+[color=embeddingcolor, mark=square*, thick, mark size=1.5pt] coordinates {
			(2048,95.49) (4096,109.35) (8192,82.95)
		};
		\end{axis}

		\setcounter{subfigure}{0}
		\node[below=4mm, font=\scriptsize, anchor=north] at (group c1r2.south) {
			\refstepcounter{subfigure}\label{fig:parameter-sensitivity:chunk}
			(\thesubfigure) Chunk size};
		\node[below=4mm, font=\scriptsize, anchor=north] at (group c2r2.south) {
			\refstepcounter{subfigure}\label{fig:parameter-sensitivity:budget}
			(\thesubfigure) Token budget};

		\node[anchor=south] at
			($(group c1r1.north)!0.5!(group c2r1.north)+(0,1.5mm)$)
			{\pgfplotslegendfromname{parameter-sensitivity-legend}};
	\end{tikzpicture}
	\caption{\blue{Scheduling-parameter sensitivity for (a) chunk size and (b) token
		budget. Throughput and latency for embedding and generation remain stable across different chunk sizes and token budgets.}}
	\label{fig:parameter-sensitivity}
\end{figure}
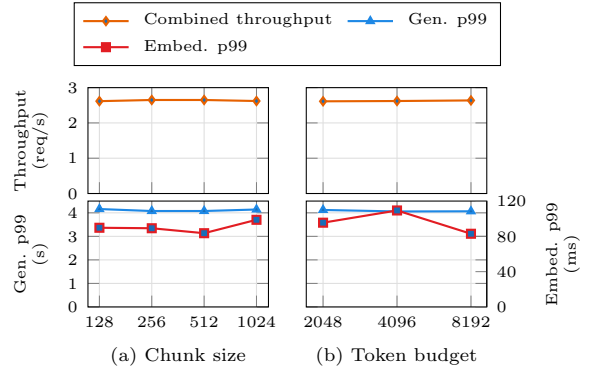

\paragraph{Chunk-Size Sweep}
\blue{We test whether performance depends on the granularity of chunked
embedding and select a chunk size for the subsequent budget sweep. Using the
5:5 workload from \cref{sec:eval:throughput}, we vary the embedding chunk size
$C$ across 128, 256, 512, and 1024 tokens. As shown in
\cref{fig:parameter-sensitivity}(a), combined throughput ranges from 2.6167 to
2.6500 requests/s, generation p99 from 4.08 to 4.16\,s, and embedding p99 from
83.58 to 98.77\,ms. This shows that performance is insensitive to chunk size over the
tested range.}

\paragraph{Token-Budget Sweep}
\blue{We next test whether performance at the selected chunk size depends on
the per-iteration token budget. The 256- and 512-token settings tie for the
highest combined throughput, so we select $C=512$ because it has the lower
embedding p99. Holding this chunk size and
the workload fixed, we sweep the token budget $B$ across 2048, 4096, and 8192
tokens. As shown in
\cref{fig:parameter-sensitivity}(b), combined throughput remains between 2.6125
and 2.6375 requests/s, generation p99 between 4.06 and 4.12\,s, and embedding
p99 between 82.95 and 109.35\,ms. Combined throughput varies by less than 1\%,
and latency shows no consistent degradation as the budget changes. These
results show that \system does not require a narrowly tuned token budget.}

\subsection{Standalone Chunked Embedding Ablation}
\label{sec:app:incremental-ablation}

\blue{Because chunked embedding and incremental pooling are enabled together
in our implementation, we evaluate their combined effect in two complementary
settings.}

\paragraph{Scheduler-Packing Simulation}
\blue{We first test whether chunking improves token-budget packing without
reducing generation service. We compare two otherwise identical FCFS iteration
schedulers. In the atomic condition, an embedding request must be processed in
one iteration. In the chunked condition, it can be divided into chunks of at
most $C=64$ tokens. Each iteration has a token budget of $B=256$, and the
request mix is 80\% generation and 20\% embedding. Each generation request has
a 128-token prefill followed by a 256-token decode, which the simulator advances
in eight-token steps. Embedding requests contain either 96 or 224 tokens, with
224-token inputs comprising 45\% of the embedding workload. As shown in
\cref{tab:incremental-simulation}, chunking raises average token-budget fill
from 8.8\% to 85.7\%. Generation throughput remains unchanged at 93.3
requests/s, while embedding throughput increases from 25.5 to 26.9 requests/s.
Thus, chunking recovers unused token capacity without sacrificing generation
throughput.}

\begin{table}[th]
	\centering
	\caption{Atomic-versus-chunked embedding simulation. Budget fill is
			the average fraction of the per-iteration token budget occupied by
			scheduled tokens; throughput is in requests/s.}
	\label{tab:incremental-simulation}
	\footnotesize
	\setlength{\tabcolsep}{3.0pt}
	\begin{tabular}{l r r r r}
		\toprule
		Embedding & \makecell{Budget\\fill (\%)} & Total & Gen & Embed \\
		\midrule
		Atomic  & 8.8  & 118.8 & 93.3 & 25.5 \\
		Chunked & 85.7 & 120.2 & 93.3 & 26.9 \\
		\bottomrule
	\end{tabular}
\end{table}

\paragraph{Embedding-Only Execution}
\blue{We next test whether the implemented chunked path introduces execution
overhead when embeddings run without generation contention. We compare
chunking disabled against chunking with incremental pooling enabled on 250
inputs using mean-pooled OPT-1.3B embeddings. As shown in
\cref{tab:incremental-standalone}, chunking reduces mean latency by 9.8\%
(20.4 to 18.4\,ms) and increases throughput by 10.9\% (49.0 to 54.4
embeddings/s). Thus, the scheduling flexibility provided by chunking does not
come at the cost of standalone embedding efficiency on this workload.}

\begin{table}[th]
	\centering
	\caption{Embedding-only comparison using mean-pooled OPT-1.3B
			embeddings on 250 inputs.}
	\label{tab:incremental-standalone}
	\footnotesize
	\setlength{\tabcolsep}{3.0pt}
	\begin{tabular}{l r r r}
		\toprule
		Setting & Avg. latency (ms) & Embed./s & Relative \\
		\midrule
		No chunk & 20.4 & 49.044 & $1.000\times$ \\
		Chunked  & 18.4 & 54.384 & $1.109\times$ \\
		\bottomrule
	\end{tabular}
\end{table}

\subsection{Latency Robustness}
\label{sec:app:latency-robustness}

\begin{figure}[t]
	\centering
	\begin{tikzpicture}
		\begin{axis}[
			name=embedding-lower-axis,
			scale only axis,
			width=0.66\linewidth,
			height=1.75cm,
			axis y line*=right,
			axis x line=none,
			label style={font=\scriptsize,align=center},
			tick label style={font=\scriptsize},
			xmin=0,
			xmax=100,
			ymin=0,
			ymax=450,
			xtick={10,50,90},
			ytick={0,150,300,450},
			]
				\addplot+[ybar,mark=none,bar width=7pt,bar shift=-4pt,
					draw=embeddingcolor!85!black,fill=embeddingcolor!25]
				coordinates {(10,278.70) (50,293.69) (90,301.87)};
				\addplot+[ybar,mark=none,bar width=7pt,bar shift=4pt,
					draw=embeddingcolor!85!black,fill=embeddingcolor!65]
				coordinates {(10,322.59) (50,327.35) (90,1706.79)};
		\end{axis}
		\begin{axis}[
			at={(embedding-lower-axis.south west)},
			anchor=south west,
			scale only axis,
			width=0.66\linewidth,
			height=2.5108cm,
			axis y line*=left,
			axis x line*=bottom,
			xlabel={Generation request share},
			ylabel={Gen. TTFT p99 (ms)},
			label style={font=\scriptsize,align=center},
			tick label style={font=\scriptsize},
			xmin=0,
			xmax=100,
			ymin=0,
			ymax=200,
			xtick={10,50,90},
			xticklabels={10\%,50\%,90\%},
			ytick={0,50,100,150,200},
				mark options={draw=generationcolor!85!black,
					fill=generationcolor,scale=0.75},
			]
				\addplot+[mark=*,thick,color=generationcolor]
				coordinates {(10,178.00) (50,181.96) (90,173.57)};
		\end{axis}
		\begin{axis}[
			name=latency-upper-axis,
			at={(embedding-lower-axis.north west)},
			anchor=south west,
			yshift=6pt,
			scale only axis,
			width=0.66\linewidth,
			height=0.55cm,
			axis y line*=right,
			axis x line=none,
			label style={font=\scriptsize,align=center},
				tick label style={font=\scriptsize},
				yticklabel style={font=\scriptsize,xshift=2pt},
			xmin=0,
			xmax=100,
			ymin=1600,
			ymax=1800,
			xtick={10,50,90},
			xticklabels={},
			ytick={1800},
			legend style={
				font=\scriptsize,
				legend columns=3,
				column sep=0.25em,
				inner xsep=2pt,
					at={(0.5,1.6)},
					anchor=south,
				},
			]
				\addplot+[ybar,mark=none,bar width=7pt,bar shift=4pt,
					draw=embeddingcolor!85!black,fill=embeddingcolor!65,
					forget plot]
				coordinates {(90,1706.79)};
			\addlegendimage{generationcolor,thick,mark=*,
				mark options={draw=generationcolor!85!black,
					fill=generationcolor,scale=0.75}}
			\addlegendentry{TTFT p99}
			\addlegendimage{area legend,draw=embeddingcolor!85!black,
				fill=embeddingcolor!25}
			\addlegendentry{Embed. p50}
			\addlegendimage{area legend,draw=embeddingcolor!85!black,
				fill=embeddingcolor!65}
			\addlegendentry{Embed. p95}
		\end{axis}
		\node[font=\scriptsize,rotate=90]
			at ($(embedding-lower-axis.south east)!0.5!
				(latency-upper-axis.north east)+(31pt,0)$)
			{Embed. latency (ms)};
		\draw[thick]
			([xshift=-2pt,yshift=-1pt]embedding-lower-axis.north east)
			-- ([xshift=2pt,yshift=2pt]embedding-lower-axis.north east);
		\draw[thick]
			([xshift=-2pt,yshift=3pt]embedding-lower-axis.north east)
			-- ([xshift=2pt,yshift=6pt]embedding-lower-axis.north east);
	\end{tikzpicture}
	\caption{Generation and embedding latency across workload ratios.
		Generation TTFT p99 and embedding p50 remain stable, while embedding p95 increases under generation-heavy workloads.}
	\label{fig:ratio-latency}
\end{figure}
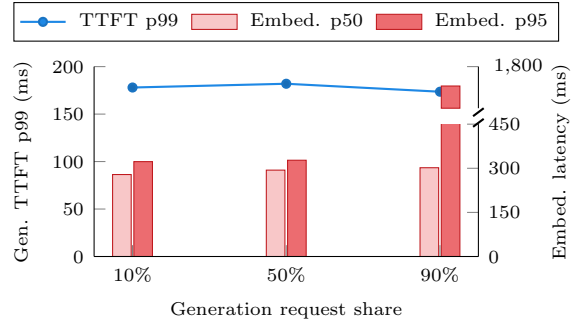

\paragraph{Workload-Ratio Sensitivity}
\blue{We evaluate whether \ibs maintains low generation startup and embedding
latency as the workload composition shifts. We sweep the generation request
share across 10\%, 50\%, and 90\%, using three repeats per setting, 512 clients,
and 180-second submission windows. Generation prompts and embedding inputs
contain 128 tokens, and generation is capped at 1024 output tokens. As shown in
\cref{fig:ratio-latency}, generation TTFT p99 remains below 200\,ms across the
sweep, ranging from 173.57 to 181.96\,ms. Embedding p50 also remains stable at
278.70--301.87\,ms. Embedding p95 is 322.59 and 327.35\,ms at 10\% and 50\%
generation, respectively, but rises to 1.71\,s at 90\% generation. Thus,
generation startup and typical embedding latency are robust to the request
mix, while an extremely generation-heavy workload primarily affects the
embedding tail.}

\paragraph{Long-Decode Sensitivity}
\blue{We evaluate whether \ibs preserves prompt embedding service as longer
generation decodes occupy the system. Using the 5:5
workload, we vary only the generation decode length, from 256 to 4096 tokens.
As shown in \cref{tab:decode-length}, generation p99 latency increases from
7.06 to 79.68\,s. In contrast, embedding p50 remains between 52.58 and
54.62\,ms, and embedding p99 remains between 53.09 and 59.92\,ms. Throughput
is not monotonic. For example, it increases from 0.1167 requests/s at 2048
tokens to 0.1333 requests/s at 4096 tokens. Thus, we do not interpret differences
between adjacent settings as a scaling trend. The central result is latency
isolation. Despite an over $11\times$ increase in generation p99, embedding p99
varies by less than 7\,ms. \ibs continues to schedule embedding requests
promptly even when generation becomes decode-dominated.}

\begin{table}[th]
	\centering
	\caption{Decode-length sensitivity on the 5:5 workload. Throughput
		is reported in requests/s and latency in milliseconds.}
	\label{tab:decode-length}
	\scriptsize
	\setlength{\tabcolsep}{1.8pt}
	\begin{tabular}{l|rr|rrr}
		\toprule
		\multirow{2}{*}{\makecell{Decode\\(tokens)}}
		& \multicolumn{2}{c|}{Throughput (rps)}
		& \multicolumn{3}{c}{Latency (ms)} \\
		\cmidrule(lr){2-3}\cmidrule(lr){4-6}
		& Gen & Embed & Gen p99 & Embed p50 & Embed p99 \\
		\midrule
		256  & 0.2333 & 0.2333 & 7057.75  & 53.32 & 59.02 \\
		512  & 0.1611 & 0.1611 & 13959.11 & 52.58 & 53.09 \\
		1024 & 0.1528 & 0.1528 & 27806.13 & 53.37 & 59.92 \\
		2048 & 0.1167 & 0.1167 & 56780.18 & 54.62 & 54.97 \\
		4096 & 0.1333 & 0.1333 & 79675.66 & 54.42 & 55.42 \\
		\bottomrule
	\end{tabular}
\end{table}

\end{document}